\documentclass[mlmain]{jmlr}

\usepackage{booktabs}
\usepackage{amssymb}
\usepackage{pifont}
\usepackage{wrapfig}

\jmlrvolume{}
\firstpageno{1}

\renewcommand{\b}[1]{\mathbf{#1}}
\newcommand{\X}{\mathcal{X}}
\newcommand{\Z}{\mathcal{Z}}
\newcommand{\M}{\mathcal{M}}
\newcommand{\R}{\mathbb{R}}
\newcommand{\Id}{\mathbb{I}}
\newcommand{\N}{\mathcal{N}}
\newcommand{\T}{\intercal}
\newcommand{\Hilb}{\mathcal{H}}

\newcommand{\CE}{{\text{CE}}}
\DeclareMathOperator*{\argmin}{arg\,min}
\newcommand{\inner}[2]{\langle#1,#2\rangle}
\def\SGD{SGD}
\def\RSGD{RSGD}
\def\RSGDCLS{RSGD-C}
\newcommand{\cmark}{\ding{51}}%
\newcommand{\xmark}{\ding{55}}%
\definecolor{purpleLight}{RGB}{178,144,255}
\definecolor{purpleDark}{RGB}{100,50,180}
\definecolor{greenLight}{RGB}{150,190,140}
\definecolor{greenDark}{RGB}{60,200,60}

\title[Counterfactual Explanations via Riemannian Latent Space Traversal]{Counterfactual Explanations via \\ Riemannian Latent Space Traversal}

 \author{\Name{Paraskevas Pegios}${}^{1,2}$ \Email{ppar@dtu.dk}\\
 \Name{Aasa Feragen}${}^{1,2}$  \Email{afhar@dtu.dk}\\
 \Name{Andreas Abildtrup Hansen}${}^{1}$  \Email{andab@dtu.dk}\\
 \Name{Georgios Arvanitidis}${}^{1}$  \Email{gear@dtu.dk}\\
 \addr Technical University of Denmark, Kongens Lyngby, Denmark \\
 \addr Pioneer Centre for AI, Copenhagen, Denmark
 }

\begin{document}
\maketitle

\begin{abstract}
The adoption of increasingly complex deep models has fueled an urgent need for insight into how these models make predictions. Counterfactual explanations form a powerful tool for providing actionable explanations to practitioners. Previously, counterfactual explanation methods have been designed by traversing the latent space of generative models. Yet, these latent spaces are usually greatly simplified, with most of the data distribution complexity contained in the decoder rather than the latent embedding. Thus, traversing the latent space naively without taking the nonlinear decoder into account can lead to unnatural counterfactual trajectories. We introduce counterfactual explanations obtained using a Riemannian metric pulled back via the decoder and the classifier under scrutiny. This metric encodes information about the complex geometric structure of the data and the learned representation, enabling us to obtain robust counterfactual trajectories with high fidelity, as demonstrated by our experiments in real-world tabular datasets.
\end{abstract}


\section{Introduction}
\label{sec:intro}

\begin{wrapfigure}{r}{0.42\textwidth}
\vspace{-60pt}
  \begin{center}
    \includegraphics[width=0.28\textwidth, height=0.27\textwidth]{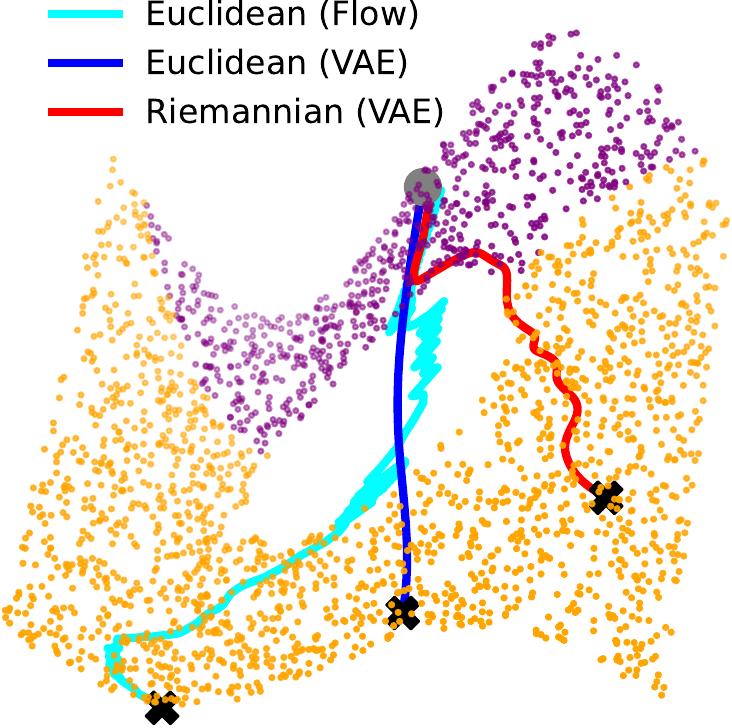}
  \end{center}
  \vspace{-25pt}
  \caption{Optimization in the latent space of generative models endowed with Euclidean metric cannot guarantee paths stay on the data manifold. Our Riemannian approach produces realistic trajectories.}
  \label{fig:teaser}
\end{wrapfigure}

Counterfactual explanations are useful to gain insights into classifier behavior, as they can be seen as answering the question \emph{``how should I transform my input data to, as efficiently as possible, change the prediction of the classifier while retaining its identity features?"}. In situations where machine learning algorithms are used to make decisions that affect our lives, counterfactual explanations could be used to give individuals feedback on how they might most easily change themselves to alter the algorithm's decision. \cite{wachter2017counterfactual} proposed to optimize the objective function, 

\begin{equation}
\label{eq:basic_cf_opt}
    \b{x}_{\CE} = \argmin_{\b{x}\in\X} \ell( c(\b{x}), y),
\end{equation}

\noindent where $\b{x}$ is the input data of interest, $y$ is the desired class, $c$ is the classifier, and $\ell$ is a loss function. However, since a simple stochastic gradient descent (SGD) optimization scheme does not take the geometry of the data manifold into account, the resulting counterfactual $\b{x}_\CE$ is not guaranteed to belong to the data manifold. As a consequence, $\b{x}_\CE$ could represent data that is very different from (i) the original input and (ii) \emph{any} realistic data point. Both issues make the counterfactual less useful for interpretation and for the ability of the individual to take action. 
Thus, a common strategy is to employ latent generative models such as Variational Autoencoders (VAEs)~\citep{kingma:iclr:2014} or Normalizing Flows~\citep{papamakarios2021jmlr} to consider counterfactuals onto the data manifold by applying~\equationref{eq:basic_cf_opt} in the latent space endowed with the standard Euclidean metric~\citep{dombrowski2023diffeomorphic}. Yet, standard decoders do not preserve the topology of the data, and thus, cannot guarantee that the counterfactual trajectories stay on the data manifold, or even lead to meaningful, smooth transitions (see \figureref{fig:teaser}).

\paragraph{We propose:} To generate counterfactual explanations via
Riemannian latent space traversal. To this end, we introduce a Riemannian geometry in the latent space, induced by a well-calibrated stochastic decoder, to respect the topology of the data manifold. By using a Riemannian Stochastic Gradient Descent (RSGD) optimizer with respect to this geometry, we ensure counterfactuals stay on the data manifold and remain realistic. On top of that, we propose a Riemannian metric which also includes the classifier under observation. Here, we aim to address potentially unstable trajectories caused by complex decision boundaries of the classifier. This allows us to create a latent geometry that not only encodes the data topology but also encourages smoother classification boundaries, yielding more realistic and actionable counterfactual trajectories. Our approach is illustrated in~\figureref{fig:model-demo}.

\section{Background and Related work}

\subsection{Riemannian latent space geometry} 
A Riemannian manifold \citep{do1992riemannian, lee2019introduction} is a well-defined metric space, where we can compute the shortest path between two points. 
A manifold $\mathcal{M}$ can be seen as a smooth $d$-dimensional surface embedded within a higher dimensional ambient space $\mathcal{X}$, for example $\mathbb{R}^D$. 
We assume that there exists a parametrization $g:\mathcal{Z}\subseteq \R^d \to\mathcal{U}$, where $\mathcal{Z}$ is the parameter space. The columns of the Jacobian $\b{J}_g(\b{z})\in\R^{D\times d}$ of $g(\cdot)$ span the tangent space $\mathcal{T}_\b{x}\mathcal{M}$ at a point $\b{x}\in\mathcal{M}$, so a tangent vector can be written as $\b{v} = \b{J}_g(\b{z})\tilde{\b{v}}$, where $\tilde{\b{v}}\in\R^d$ are the \emph{intrinsic coordinates} of the tangent vector on the associated tangent space. 

A positive definite matrix $\b{M}_{\X}:\mathcal{X}\to \mathcal{R}^{D\times D}_{\succ 0}$ that changes smoothly across the ambient space $\mathcal{X}$ is 
a Riemannian metric, and can be used to define the inner product between two tangent vectors $\inner{\b{u}}{\b{v}}_{\b{x}} = \inner{\b{u}}{\b{M}_{\X}(\b{x})\b{v}} = \inner{\tilde{\b{u}}}{\b{J}_g(\b{z})^\T\b{M}_{\X}(g(\b{z}))\b{J}_g(\b{z})\tilde{\b{v}}}$. Note that the matrix $\b{M}_{\mathcal{Z}}(\b{z}):= \b{J}_g(\b{z})^\T\b{M}_{\X}(g(\b{z}))\b{J}_g(\b{z}) \in \R^{d\times d}$ is positive definite, and also changes smoothly if $g(\cdot)$ is at least twice differentiable, hence, it is a Riemannian metric in the space $\mathcal{Z}$. This metric $\b{M}_\Z(\cdot)$ captures the geometry of $\M$, while respecting the geometry induced by $\b{M}_\X(\cdot)$, and intuitively, shortest paths prefer regions where its magnitude 
is small.

In the ambient space $\X$, we expect the given data $\{\b{x}_n\}_{n=1}^N$ to lie near a low-dimensional manifold $\M$ that has an unknown geometric structure~\citep{bengio2013representation}. We can approximate the data by considering \emph{latent representations} $\b{z}_n$ and learning a function $\b{x} \approx \hat{g}(\b{z})$. This function does not correspond to the true parametrization $g(\cdot)$ of $\M$, yet it can approximate relatively well the implicit data manifold $\M\approx \hat{g}(\Z)$, and hence, induce a \emph{pull-back} metric in $\Z$ which approximates the geometry of $\M$. In practice, $\hat{g}(\cdot)$ can be learned with deep generative models, i.e., stochastic generators \citep{hauberg:arxiv:2018,arvanitidis:iclr:2018}

\subsection{Counterfactual explanations}
Recently, different counterfactual explanation (CE) methods have been proposed~\citep{verma2020counterfactual}. Early works address key aspects of counterfactual analysis, such as sparsity~\citep{guidotti2018local,dandl2020multi}, actionability~\citep{ustun2019actionable},  diversity~\citep{russell2019efficient,mothilal2020explaining}, and causality~\citep{mahajan2019preserving,karimi2020algorithmic}.~\cite{borisov2022deep} underlines two categories of CE methods: One assuming feature independence~\citep{karimi2020model} and one focusing on feature dependencies, exploring data geometry~\citep{pawelczyk2020learning,downs2020cruds}. Note that CEs are different from counterfactual editing whose goal is \emph{not} to explain changes in classifier's predictions~\citep{melistas2024benchmarking}. When data lie on a manifold, attempts have been made to maintain CEs close to the data by constructing connective graphs~\citep{poyiadzi2020face} or employing distance measures that can identify feasible paths~\citep{domnich2024enhancing}. Other approaches approximate manifold spaces with latent generative models, including VAE-~\citep{dhurandhar2018explanations,joshi2019towards,antoran2020getting}, GAN-~\citep{singla2019explanation}, flow-~\citep{dombrowski2021diffeomorphic,duong2023ceflow}, and diffusion-based methods~\citep{jeanneret2022diffusion,weng2025fast,pegios2024diffusion,madaan2023diffusion}. In this work, we employ VAEs to induce a Riemannian latent space geometry, enabling the generation of realistic CEs with trajectories that remain on the data manifold.

\begin{figure}[t]
\centering
\floatconts
  {fig:subfigex}
  {\caption{Conceptual demonstration: a) the latent space and the two distinct metrics $\color{purpleLight} \b{M}_\Z$ and $\color{greenLight}\widehat{\b{M}}_\Z$, b) the data space with the two proposed options for the ambient metrics $\color{purpleDark}\Id_D$ and $\color{greenDark}\b{M}_\X$, and c) the representation space and the linear classifier. The ambient metric $\color{greenDark}\b{M}_\X$ in $\X$ captures the geometry of the hidden representation manifold, and we pull it back in $\Z$ to get $\color{greenLight}\widehat{\b{M}}_\Z$, while the metric $\color{purpleLight} \b{M}_\Z$ captures only the geometry of the data manifold. See Section 3 for further details.}\label{fig:model-demo}}
  {%
    \subfigure[\scriptsize Latent space and the two distinct Riemannian metrics $\color{purpleLight} \b{M}_\Z$ and $\color{greenLight}\widehat{\b{M}}_\Z$.]{\label{fig:demo1}%
      \includegraphics[height=4cm]{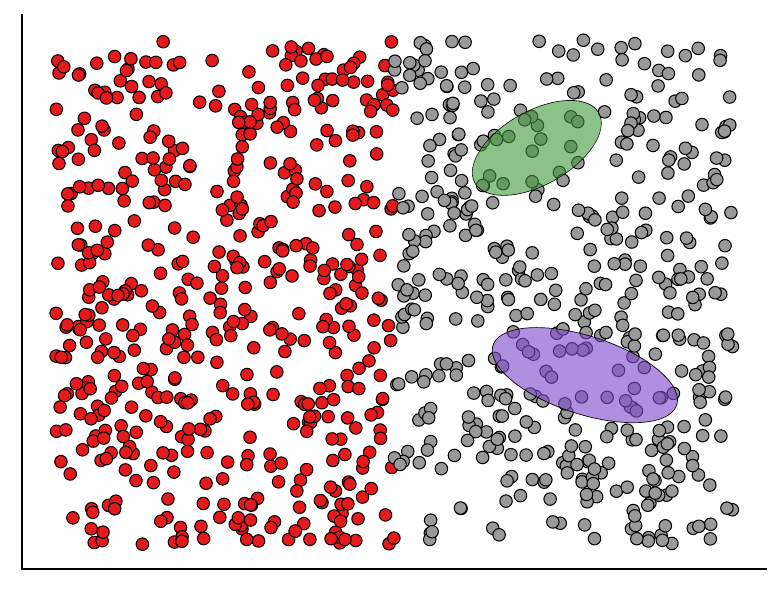}}%
    ~
    \subfigure[\scriptsize The data manifold and two ambient metrics $\color{purpleDark}\Id_D$ and $\color{greenDark}\b{M}_\X$.]{\label{fig:demo2}%
      \includegraphics[height=4cm]{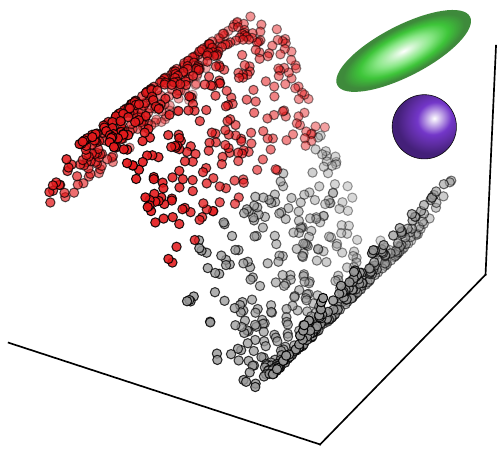}}
    ~
    \subfigure[\scriptsize Representation space manifold together with a linear classifier.]{\label{fig:demo3}%
      \includegraphics[height=4cm]{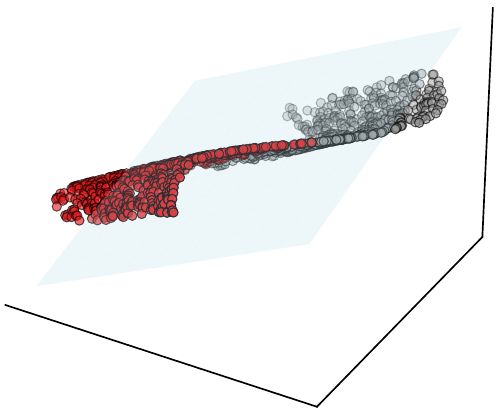}}
  }
\end{figure}

\section{Theory}
\label{sec:theory}

Let a dataset $\{\b{x}_n,y_n\}_{n=1}^N$ in an ambient space $\X=\R^D$ where $y_n\in\{0,1\}$. \cite{dombrowski2023diffeomorphic} generate CEs by applying SGD in the latent space $\Z=\R^d$ of a generative model, moving through latent space taking steps with size $\eta$ of the form: $\b{z}' = \b{z} - \eta \cdot \nabla \ell(c(\hat{g}(\b{z})), y)$. Yet, this risks ``falling off" the data manifold, forcing the generative model to extrapolate beyond its training data, resulting in unrealistic counterfactual paths (see ~\figureref{fig:teaser}).

\subsection{Basic pull-back metric}

The standard VAE is used for learning stochastic generative models with Gaussian decoders \citep{kingma:iclr:2014}. 
Specifically, we consider the stochastic decoder $\b{x} = g_\varepsilon(\b{z}) = \mu(\b{z}) + \sigma(\b{z})\odot\epsilon$, with $\epsilon\sim\mathcal{N}(0,\Id_D)$ and $\odot$ being the point-wise product. 
When the ambient space $\X$ is endowed with the Euclidean metric $\b{M}_\X(\b{x})=\Id,~\forall\b{x}\in\X$, the expected pull-back metric in the latent $\Z=\R^d$ space is equal to
\begin{equation}
\label{eq:gaussian_pullback}
    \b{M}_\Z(\b{z}) = \mathbb{E}_\varepsilon[\b{J}_{g_\varepsilon}(\b{z})^\T\b{J}_{g_\varepsilon}(\b{z})] = \b{J}_\mu^\intercal(\b{z})\b{J}_\mu(\b{z}) +\b{J}_\sigma(\b{z})^\T \b{J}_\sigma(\b{z})
\end{equation}
where $\mu \colon \Z \to \X$ and $\sigma \colon \Z \to \R_{>0}$ are neural networks \citep{arvanitidis:iclr:2018}. We induce a meaningful geometric structure in the latent space whether the generator's uncertainty is well-calibrated, i.e, $\sigma(\b{z})\to 0$ when $\b{z}$ is near the training latent codes, otherwise $\sigma(\b{z})\to +\infty$ \citep{hauberg:arxiv:2018}.
We ensure a well-calibrated uncertainty by using $\sigma(\b{z})={1}/{\sqrt{\gamma(\b{z})}}$, with $\gamma \colon \Z \to \R_{>0}$ a Radial Basis Function network~\citep{que2016back} based on the latent representation, satisfying the criteria above~\citep{arvanitidis:iclr:2018}.

Under the mild conditions on the smoothness of $g_\varepsilon(\cdot)$ and the behavior of $\sigma(\cdot)$, $\b{M}_\Z(\b{z})\in\R^{D\times D}_{\succ 0}$ constitutes a Riemannian metric in $\Z$, which captures the geometry of the data manifold in the ambient space $\M\subset \X$, and hence the shortest paths between points in $\Z$ respect this geometry. Therefore, as the cost of moving off the data manifold increases with the VAE uncertainty, the result is the corresponding shortest paths to prefer staying near the data manifold. Thus, the topology of the data manifold is preserved in a soft sense: It is \emph{possible}, but \emph{expensive}, for shortest paths, and hence counterfactual trajectories to move outside the data manifold, meaning they do not do so unless necessary. This allows us to generate CEs using RSGD w.r.t. the pull-back metric defined in~\equationref{eq:gaussian_pullback},

\begin{equation}
    \label{eq:riem_gd_basic}
    \b{z}' = \b{z} - \eta \cdot \frac{\b{r}}{\| \b{r}\|_{2}}, \quad \text{where}\quad \b{r} = \b{M}_\Z^{-1}(\b{z}) \nabla_{\b{z}} f_y(\b{z}) \quad \text{and} \quad f_y(\b{z}) = \ell( c(\mathbb{E}[g_\varepsilon(\b{z})]), y).
\end{equation}
We normalize the Riemannian gradient $\b{r}$ as we are interested in the update direction, and to avoid vanishing gradients near the boundary of the latent data support in $\Z$. Note that this Riemannian metric makes the optimization trajectory invariant under reparametrization, i.e., another latent representation that generates the same density in $\X$ will provide the same trajectory on the data manifold. Hence, using RSGD, we encourage trajectories to align with the geometric structure of the latent codes, staying on the data manifold.

\subsection{Enhanced pull-back metric}
\equationref{eq:gaussian_pullback} assumes a Euclidean metric on the ambient space $\X$, but we are able to consider other Riemannian metrics on $\X$ to better capture the desired properties of a problem, enriching the latent space geometry with additional information \citep{arvanitidis:aistats:2021}.

To this end, we consider a differentiable ambient space classifier $c\colon \X\to[0,1]$, which we parameterize with a neural network, and thus can be expressed as $c(\b{x}) = \text{sigmoid}(\b{w}^\T h(\b{x}))$, where $h\colon \X\to\Hilb=\R^H$ with typically $H\gg D$, and $\b{w}\in\R^H$ the weights of the last layer. 
The mapping $h(\cdot)$ is the learned representation of the final layer, spanning a $D$-dimensional submanifold of $\R^H$ under mild conditions. By endowing $\Hilb$, where the classification boundary is linear, with an Euclidean metric, we derive the pull-back Riemannian metric in $\X$,

\begin{equation}
\label{eq:representation_metric}
    \b{M}_\X(\b{x}) = \b{J}_h(\b{x})^\T \b{J}_h(\b{x}).
\end{equation}
This metric represents in $\X$ the corresponding local geometry of the learned representation $\Hilb$. Assuming well-separated classes in $\Hilb$, we expect $\b{M}_\X(\b{x})$ to be small in regions of $\X$ where the classes remain constant and to increase near the decision boundary. In principle, ~\equationref{eq:representation_metric} could used for counterfactual optimization directly in $\X$ that respects the geometry of the learned representation. Yet, as $\X$ is typically high-dimensional, it is more advantageous to exploit a generative model and pull $\b{M}_\X(\b{x})$ back to the associated latent space $\Z$ using the stochastic generator $\b{x} = g_\varepsilon(\b{z})$ to get the Riemannian metric, 
\begin{equation}
    \b{M}^\varepsilon_{\Z,\X}(\b{z}) = \b{J}_{g_\varepsilon}(\b{z})^\T \b{M}_\X(g_\varepsilon(\b{z})) \b{J}_{g_\varepsilon}(\b{z}),
    \label{eq:pullback_representation}
\end{equation}
which is stochastic due to the random variable $\varepsilon$. \figureref{fig:model-demo} illustrates our approach using the classifier-guided enhanced pull-back metric. For practical purposes, we use this metric
\begin{equation}
    \widehat{\b{M}}_\Z(\b{z})=\mathbb{E}_\varepsilon[\b{M}^\varepsilon_{\Z,\X}(\b{z})] \approx \b{J}_\mu(\b{z})^\T\b{M}_\X(\mu(\b{z}))\b{J}_\mu(\b{z}) + \b{J}_\sigma(\b{z})^\T\b{M}_\X(\mu(\b{z}))\b{J}_\sigma(\b{z}).
\end{equation}
which is a good approximation of the expected metric for well-calibrated decoders~\citep{arvanitidis:aistats:2021}. We perform CE optimization~(\equationref{eq:riem_gd_basic}) using RSGD w.r.t to the classifier-guided enhanced pull-back metric $\widehat{\b{M}}_\Z(\b{z})$, and refer to it as RSDG-C.

\section{Experiments} \label{sec:exp}
First, we show that, unlike Normalizing Flows~\citep{dombrowski2021diffeomorphic}, our induced geometry softly preserves the dataset topology by heavily penalizing counterfactual trajectories through data-sparse regions. Next, we compare Euclidean \SGD{} on VAE's latent space with our proposed Riemannian optimizers, \RSGD{} and \RSGDCLS{}, on real-world tabular datasets. Our source code is publicly available at \url{https://github.com/ppegiosk/RSGD-C}.

\begin{figure}[t]
\centering
\floatconts
  {fig:subfigex}
  {\caption{Optimization paths in the latent space of VAE using our proposed metric.}\label{fig:model-example}}
  {%
    \subfigure[\scriptsize Proposed Metric]{\label{fig:proposed-metric}%
      \includegraphics[width=0.3\linewidth]{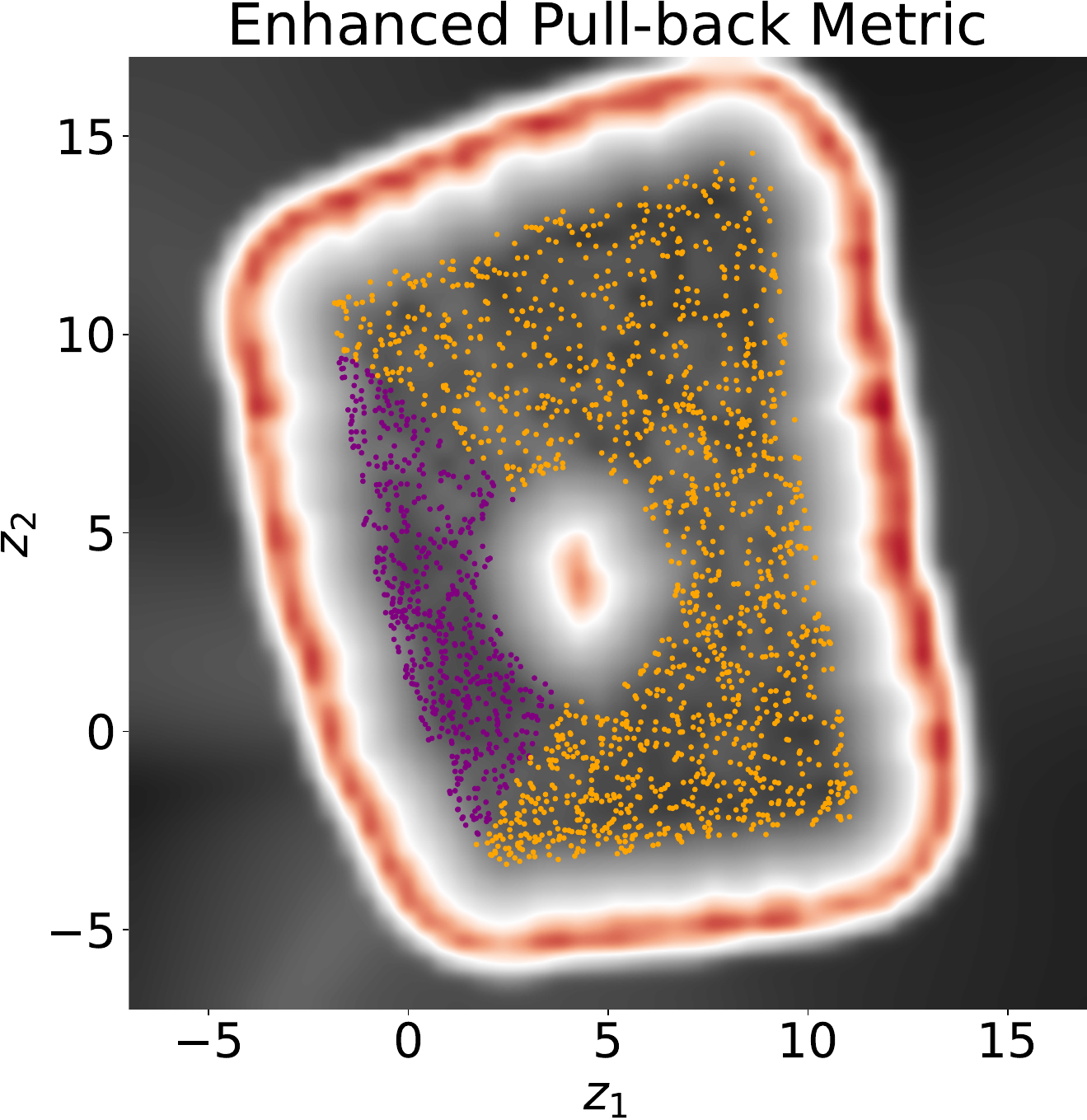}}%
    \quad
    \subfigure[\scriptsize Latent space]{\label{fig:opt-latent}%
      \includegraphics[width=0.3\linewidth]{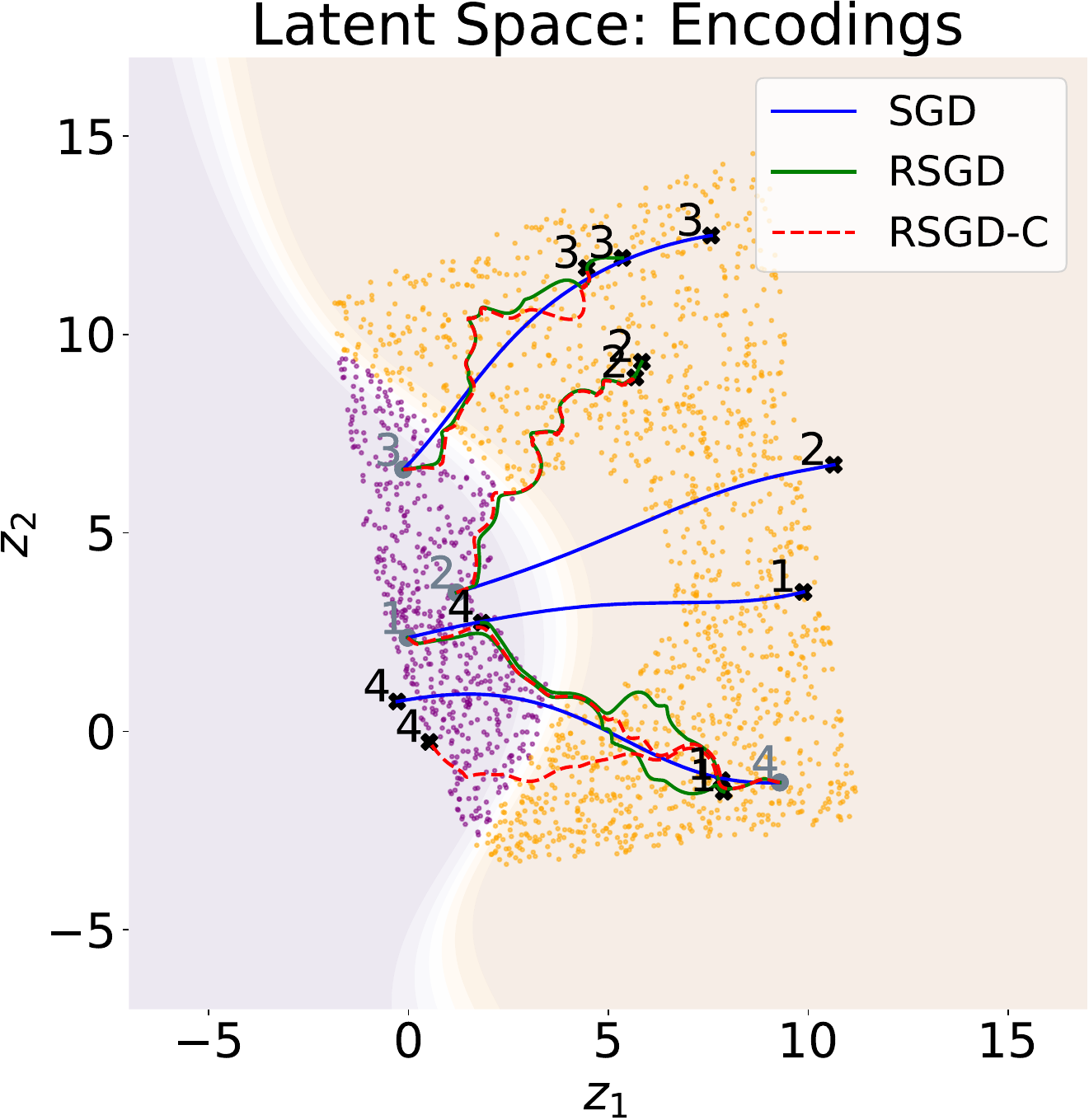}}
       \quad
    \subfigure[\scriptsize CE trajectories]{\label{fig:opt-paths}%
      \includegraphics[width=0.3\linewidth]{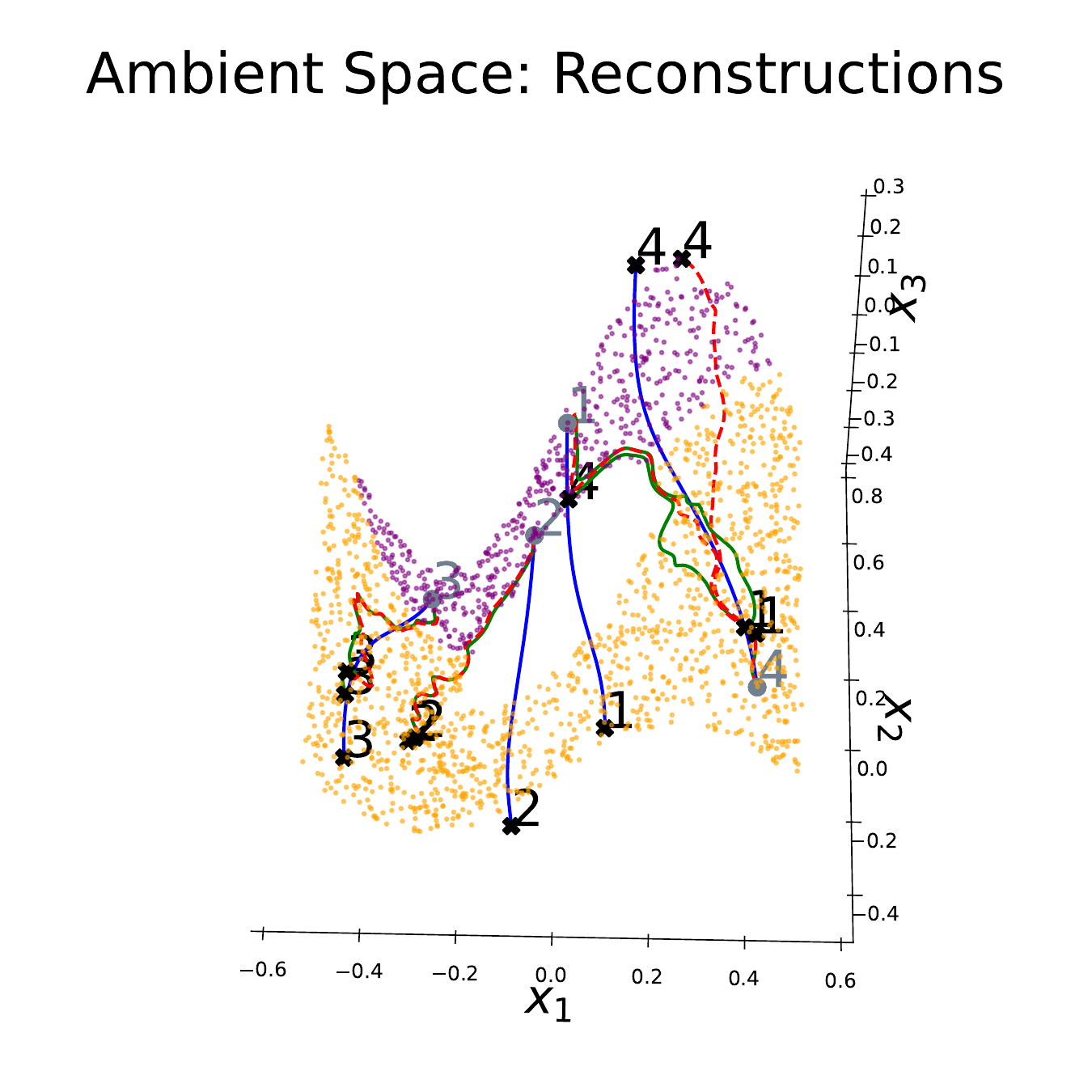}}
  }
\end{figure}

\begin{figure}[t]
\centering
\floatconts
  {fig:subfigex}
  {\caption{Optimization paths in the latent space of Normalizing Flow.}\label{fig:flow}}
  {%
    \subfigure[\scriptsize Samples from prior]{\label{fig:flow-samples}%
      \includegraphics[width=0.3\linewidth]{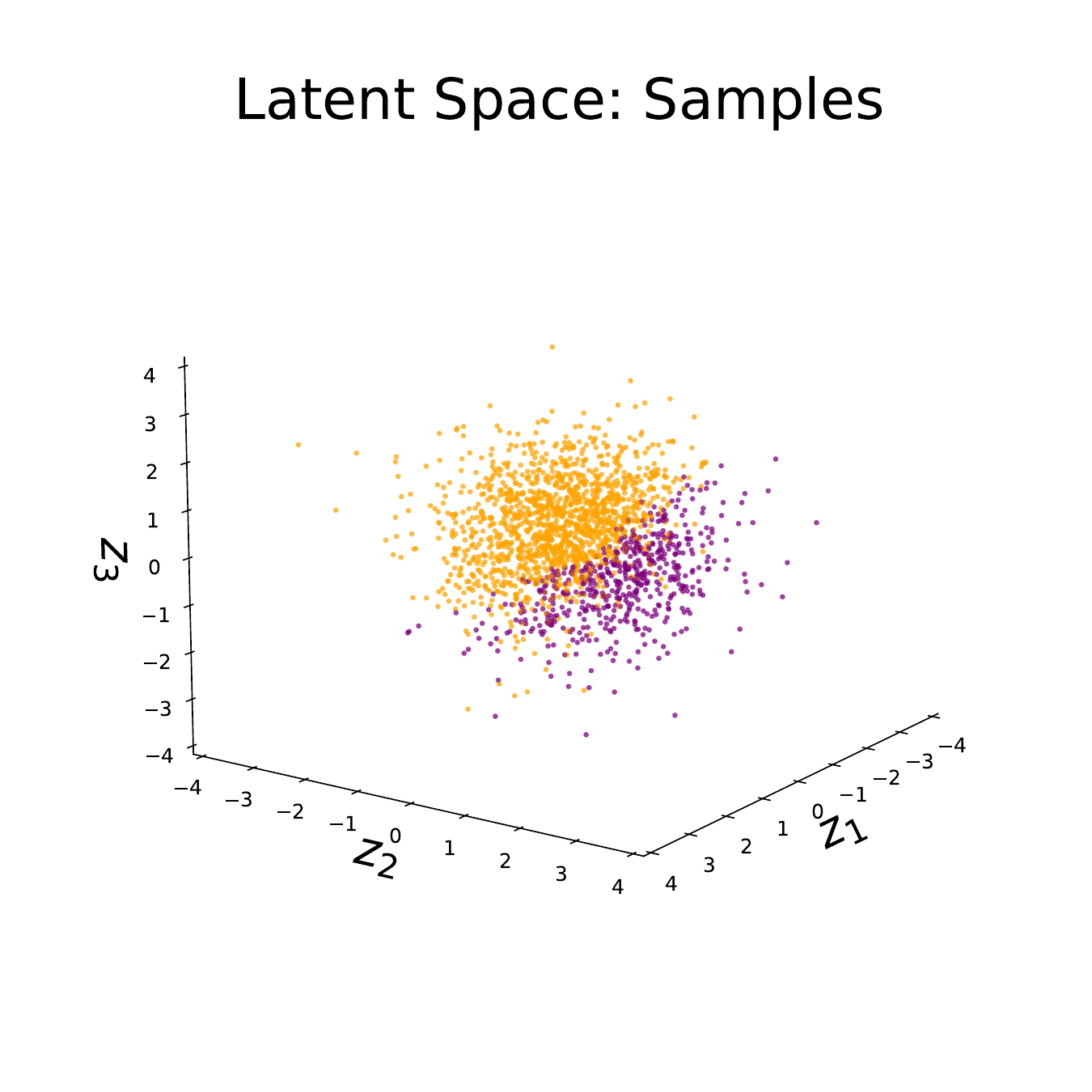}}%
    \quad
    \subfigure[\scriptsize Latent space]{\label{fig:flow-encode}%
      \includegraphics[width=0.3\linewidth]{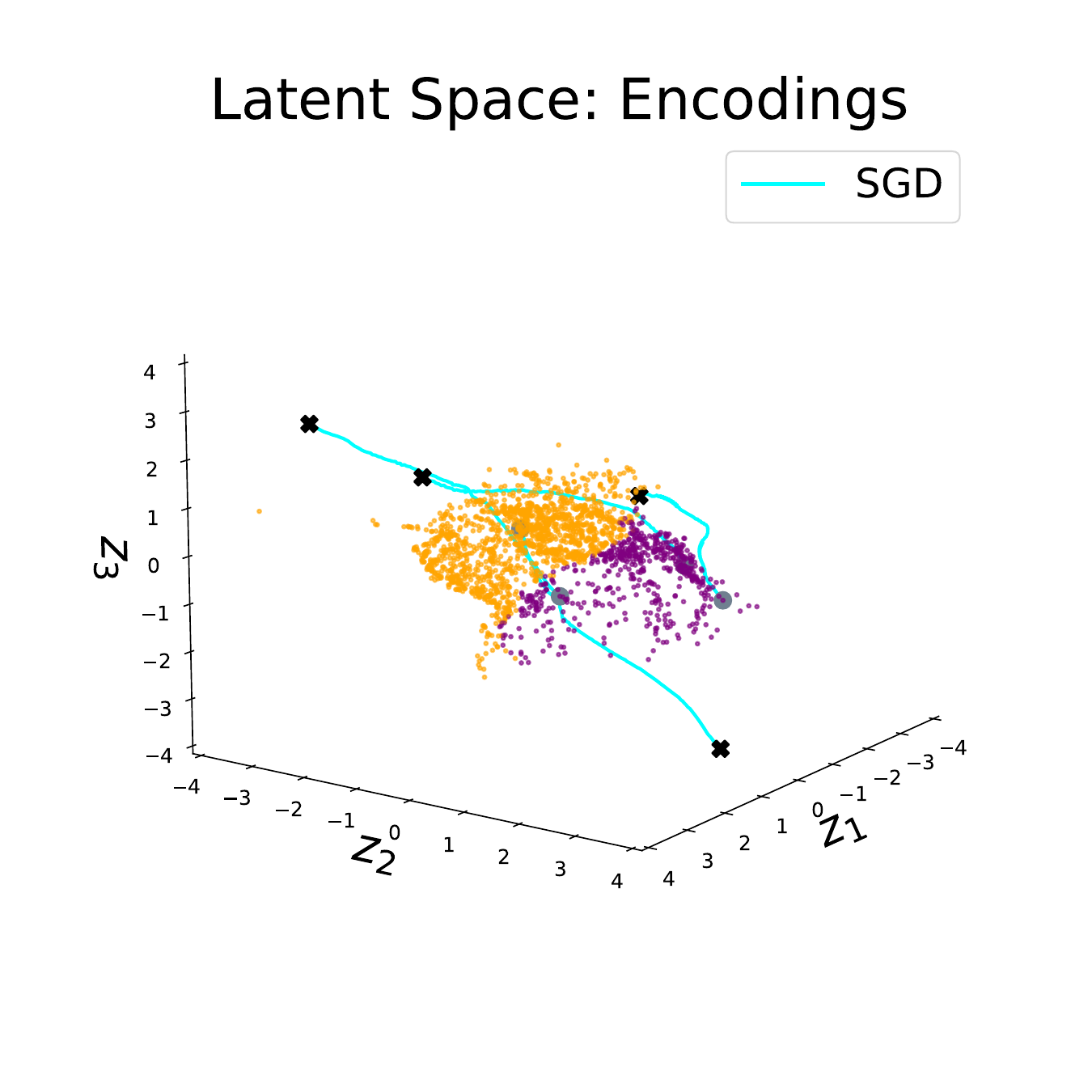}}
       \quad
    \subfigure[\scriptsize CE trajectories]{\label{fig:flow-ambient}%
      \includegraphics[width=0.3\linewidth]{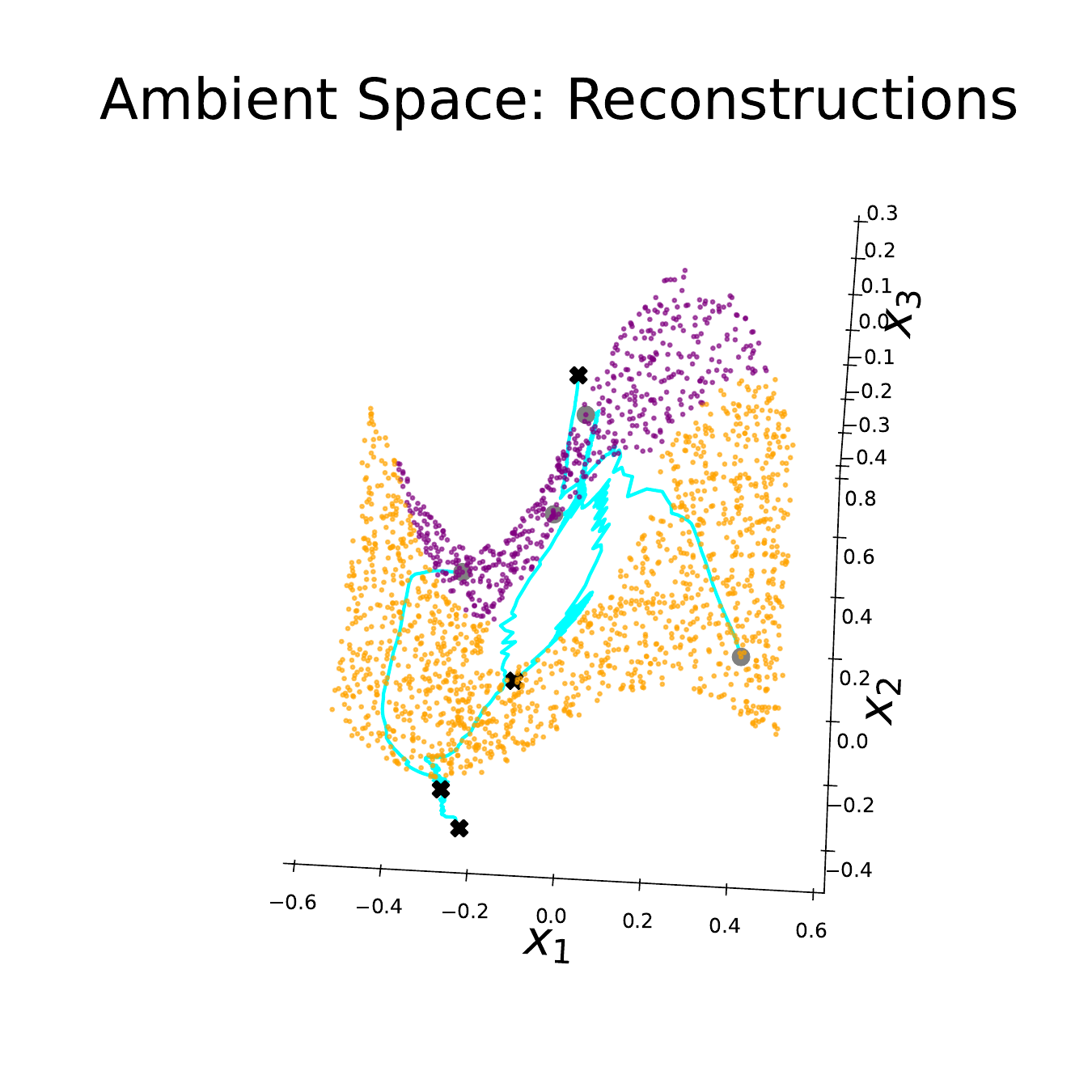}}
  }
\end{figure}

\subsection{Topology preservation and counterfactual trajectories}

We construct a surface in $\X=\R^3$ as $\b{x}=[\b{z}, 0.25 \cdot \sin(z_1)] + \varepsilon$ where ${z}_j\sim\mathcal{U}(0,2\pi),~j=1,2$ and $\varepsilon\sim \N(0, 0.1^2 \cdot \Id_3)$, with a hole in the middle by removing the points in the center with radius $||\b{z}||_2 < 0.2$ before the mapping in $\R^3$, and assign labels $y = (\text{sign}(z_2 - 2.5 \cdot z_1^2) + 1) \cdot 0.5$.
We train a Normalizing Flow with a latent space of $d=D=3$, a VAE with $d=2$, and a classifier with a representation space of $H=32$ to pull geometry back to the latent space. 

The proposed metric is illustrated in~\figureref{fig:proposed-metric}, with a black-white-red color scheme representing traversal costs in the latent space, from low to high. The red area in the center indicates the high cost of crossing the data-sparse hole, while the white-and-red boundary highlights the high cost of leaving the data-populated region. \figureref{fig:opt-latent,fig:opt-paths} show, from the viewpoint of the latent space and ambient space, respectively, that naive Euclidean latent space traversal using standard SGD fails to respect the data manifold’s topology as it is neither encoded by the classifier nor the latent space geometry, allowing counterfactuals to stray from the data distribution. In contrast, using pullback metrics from the ambient space (\RSGD{}) or the target of the classifier (\RSGDCLS{}) addresses this by making it costly for counterfactual trajectories to deviate from the data. Next, \figureref{fig:flow-samples} displays the prior distribution of trained Normalizing Flow, which poorly represents the hole. \figureref{fig:flow-encode,fig:flow-ambient} show its trajectories in the latent and ambient spaces, respectively, revealing their failure to capture data topology resulting in paths that pass through the hole.

\subsection{Improved fidelity of counterfactual trajectories}

\subsubsection{Datasets}
We evaluate our optimizers, against standard Euclidian \SGD{} on two widely used tabular datasets. \textbf{Adult}~\citep{misc_adult_2} includes census records with $D=13$ mixed-type features, intending to predict whether an individual's income exceeds \$50K/year. \textbf{Give Me Some Credit} (GMC)~\citep{Kaggle2011} includes $D=10$ continuous features, aiming to predict whether an individual will face financial distress. Both datasets include a subset of features that are considered immutable, e.g., age, as altering them cannot provide actionable feedback. More information can be found in the Appendix.

\subsubsection{Evaluation Metrics}
We follow the evaluation protocol of~\cite{pawelczyk1carla}, by measuring \emph{violation}, i.e, how often a CE method breaches user-defined constraints (immutable features), as well as \emph{validity} with Flip Ratio (FR), representing the success rate of CEs classified as the target label, along with their confidence $c(\b{x}_{\CE})$. The quality of valid CEs is measured by their \emph{closeness} to the original factual input using $\mathcal{L}_0$, $\mathcal{L}_1$, $\mathcal{L}_2$ and $\mathcal{L}_{\infty}$ distances. To measure \emph{realism}, i.e., how close the CE is to the data manifold, we compute the local Euclidean distance of CE to the closest training data point in the ambient space denoted as $\mathcal{L}_{D}$. 

\begin{table}[t]
    \centering
    \scriptsize
    
\resizebox{\linewidth}{!}{
\begin{tabular}{c|c|c||ccccc|ccc} \toprule
        \#Iter & Constrains  & Optimizer & $\mathcal{L}_{D} \downarrow$ & $\mathcal{L}_{0} \downarrow$ & $\mathcal{L}_{1} \downarrow$  & $\mathcal{L}_{2} \downarrow$ &  $\mathcal{L}_{\infty} \downarrow$ & $c(\b{x}_{\CE}) \uparrow$ & FR $\uparrow$ & violation $\downarrow$\\ \midrule
 &  & \SGD & 0.20\tiny{$\pm$0.29} & 6.86\tiny{$\pm$0.83} & 2.97\tiny{$\pm$0.98} & 2.22\tiny{$\pm$0.97} & 0.95\tiny{$\pm$0.11} & \textbf{0.93\tiny{$\pm$0.07}} & \textbf{0.99} & 1.06\\
50 & \xmark & \RSGD & \textbf{0.04\tiny{$\pm$0.05}} & \textbf{6.01\tiny{$\pm$0.08}} & \textbf{1.00\tiny{$\pm$0.43}} & \textbf{0.40\tiny{$\pm$0.36}} & 0.48\tiny{$\pm$0.24} & 0.66\tiny{$\pm$0.22} & 0.70 & \textbf{0.97}\\ 
 &  & \RSGDCLS & \textbf{0.04\tiny{$\pm$0.07}} & 6.12\tiny{$\pm$0.33} & 1.03\tiny{$\pm$0.52} & 0.42\tiny{$\pm$0.46} & \textbf{0.47\tiny{$\pm$0.25}} & 0.63\tiny{$\pm$0.21} & 0.67 & \textbf{0.97}\\ \cmidrule{2-11}

 &  & \SGD & 0.12\tiny{$\pm$0.14} & 6.26\tiny{$\pm$0.54} & 1.60\tiny{$\pm$0.73} & 0.87\tiny{$\pm$0.68} & 0.69\tiny{$\pm$0.21} & \textbf{0.91\tiny{$\pm$0.07}} & \textbf{0.99} & 1.00\\
50 & \cmark & \RSGD & \textbf{0.05\tiny{$\pm$0.06}} & \textbf{6.01\tiny{$\pm$0.09}} & \textbf{0.91\tiny{$\pm$0.34}} & \textbf{0.31\tiny{$\pm$0.24}} & \textbf{0.43\tiny{$\pm$0.17}} & 0.66\tiny{$\pm$0.21} & 0.71 & \textbf{0.96}\\ 
 & & \RSGDCLS & \textbf{0.05\tiny{$\pm$0.06}} & 6.09\tiny{$\pm$0.30} & 0.94\tiny{$\pm$0.45} & 0.35\tiny{$\pm$0.37} & 0.44\tiny{$\pm$0.22} & 0.63\tiny{$\pm$0.21} & 0.67 & \textbf{0.96}\\ \midrule \midrule

 &  & \SGD & 0.33\tiny{$\pm$0.38} & 7.08\tiny{$\pm$1.07} & 3.37\tiny{$\pm$1.21} & 2.58\tiny{$\pm$1.21} & 0.96\tiny{$\pm$0.10} & \textbf{0.94\tiny{$\pm$0.06}} & \textbf{0.99} & 1.10\\
100 & \xmark & \RSGD & \textbf{0.06\tiny{$\pm$0.06}} & \textbf{6.02\tiny{$\pm$0.17}} & 1.44\tiny{$\pm$0.52} & 0.79\tiny{$\pm$0.50} & 0.70\tiny{$\pm$0.27} & 0.85\tiny{$\pm$0.14} & 0.96 & \textbf{0.98}\\ 
 &  & \RSGDCLS & 0.07\tiny{$\pm$0.07} & 6.14\tiny{$\pm$0.37} & \textbf{1.41\tiny{$\pm$0.61}} & \textbf{0.73\tiny{$\pm$0.60}} & \textbf{0.63\tiny{$\pm$0.27}} & 0.83\tiny{$\pm$0.14} & 0.95 & \textbf{0.98}\\ \cmidrule{2-11}

 &  & \SGD & 0.13\tiny{$\pm$0.14} & 6.37\tiny{$\pm$0.66} & 1.72\tiny{$\pm$0.85} & 0.98\tiny{$\pm$0.80} & 0.71\tiny{$\pm$0.22} & \textbf{0.92\tiny{$\pm$0.06}} & \textbf{0.99} & 1.01\\
100 & \cmark & \RSGD & \textbf{0.08\tiny{$\pm$0.06}} & \textbf{6.03\tiny{$\pm$0.18}} & \textbf{1.17\tiny{$\pm$0.38}} & \textbf{0.49\tiny{$\pm$0.33}} & 0.55\tiny{$\pm$0.18} & 0.85\tiny{$\pm$0.14} & 0.95 & \textbf{0.98}\\ 
 &  & \RSGDCLS & 0.09\tiny{$\pm$0.07} & 6.13\tiny{$\pm$0.35} & 1.20\tiny{$\pm$0.52} & 0.52\tiny{$\pm$0.48} & \textbf{0.53\tiny{$\pm$0.22}} & 0.83\tiny{$\pm$0.14} & 0.95 & \textbf{0.98}\\ \midrule \midrule

 &  & \SGD & 0.41\tiny{$\pm$0.49} & 7.26\tiny{$\pm$1.21} & 3.65\tiny{$\pm$1.44} & 2.84\tiny{$\pm$1.42} & 0.96\tiny{$\pm$0.10} & \textbf{0.94\tiny{$\pm$0.06}} & \textbf{0.99} & 1.14\\
150 & \xmark & \RSGD & \textbf{0.05\tiny{$\pm$0.07}} & \textbf{6.07\tiny{$\pm$0.29}} & \textbf{1.80\tiny{$\pm$0.47}} & \textbf{1.14\tiny{$\pm$0.44}} & 0.88\tiny{$\pm$0.17} & 0.89\tiny{$\pm$0.13} & 0.97 & 0.99\\ 
 & & \RSGDCLS & \textbf{0.05\tiny{$\pm$0.08}} & 6.20\tiny{$\pm$0.44} & 1.83\tiny{$\pm$0.61} & 1.15\tiny{$\pm$0.62} & \textbf{0.83\tiny{$\pm$0.20}} & 0.89\tiny{$\pm$0.12} & 0.97 & 0.99\\ \cmidrule{2-11}

 &  & \SGD & 0.14\tiny{$\pm$0.15} & 6.42\tiny{$\pm$0.71} & 1.79\tiny{$\pm$0.90} & 1.04\tiny{$\pm$0.84} & 0.72\tiny{$\pm$0.22} & \textbf{0.92\tiny{$\pm$0.06}} & \textbf{0.99} & 1.02\\
150 & \cmark & \RSGD & \textbf{0.09\tiny{$\pm$0.07}} & \textbf{6.09\tiny{$\pm$0.32}} & \textbf{1.36\tiny{$\pm$0.45}} & \textbf{0.65\tiny{$\pm$0.44}} & \textbf{0.63\tiny{$\pm$0.19}} & 0.88\tiny{$\pm$0.13} & 0.97 & \textbf{0.98}\\ 
 &  & \RSGDCLS & \textbf{0.09\tiny{$\pm$0.08}} & 6.22\tiny{$\pm$0.40} & 1.46\tiny{$\pm$0.58} & 0.77\tiny{$\pm$0.59} & 0.66\tiny{$\pm$0.21} & 0.88\tiny{$\pm$0.12} & 0.97 & \textbf{0.98}\\ \bottomrule

    \end{tabular}
    }
    \caption{Adult dataset. To standardize comparison, we evaluate optimizers with (\cmark) and without (\xmark) fidelity constraints \emph{during} optimization for different numbers of steps.}
    \label{tab:results-adult}
\end{table}

\begin{figure}[t]
\centering

  {%
    {\label{fig:ld-noconstrains}%
      \includegraphics[width=0.35\linewidth]{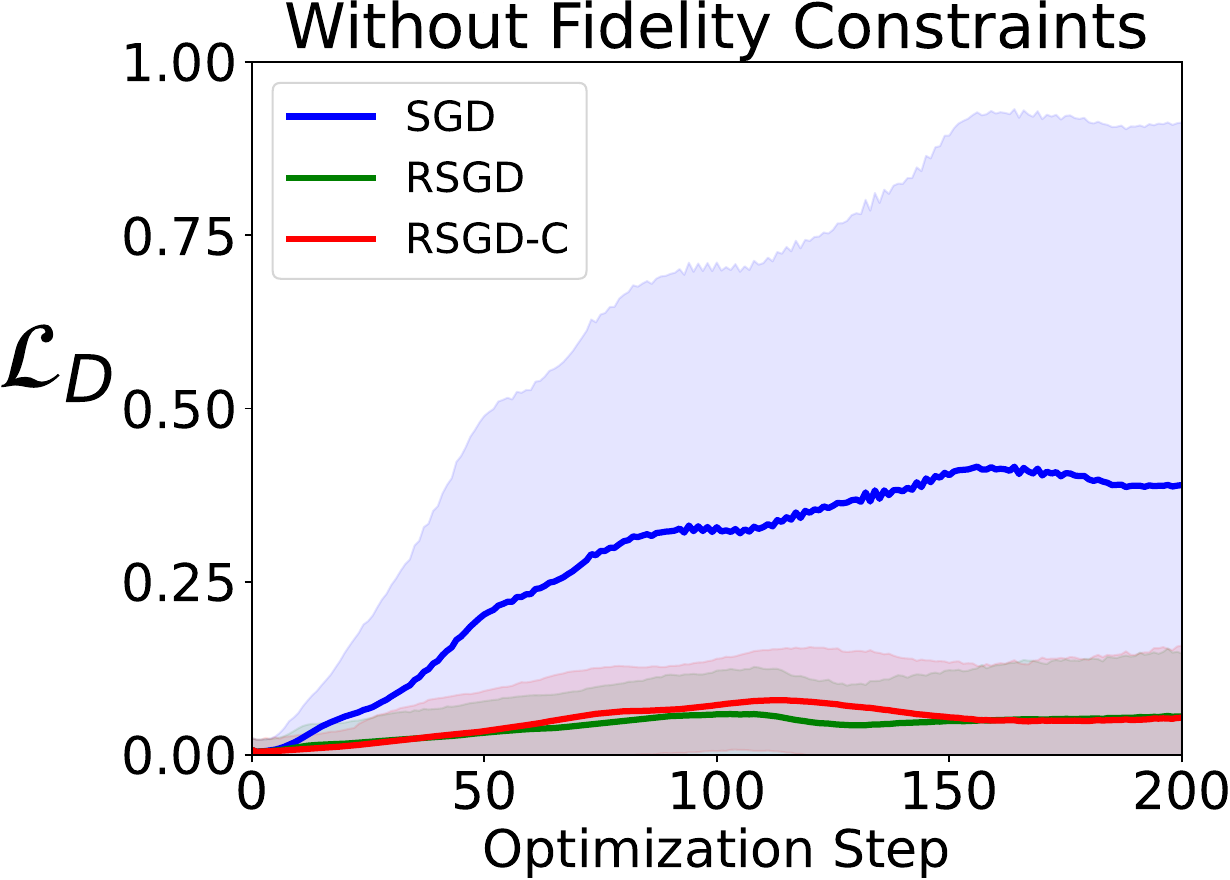}}%
    \qquad \qquad
    {\label{fig:ld-constrains}%
      \includegraphics[width=0.35\linewidth]{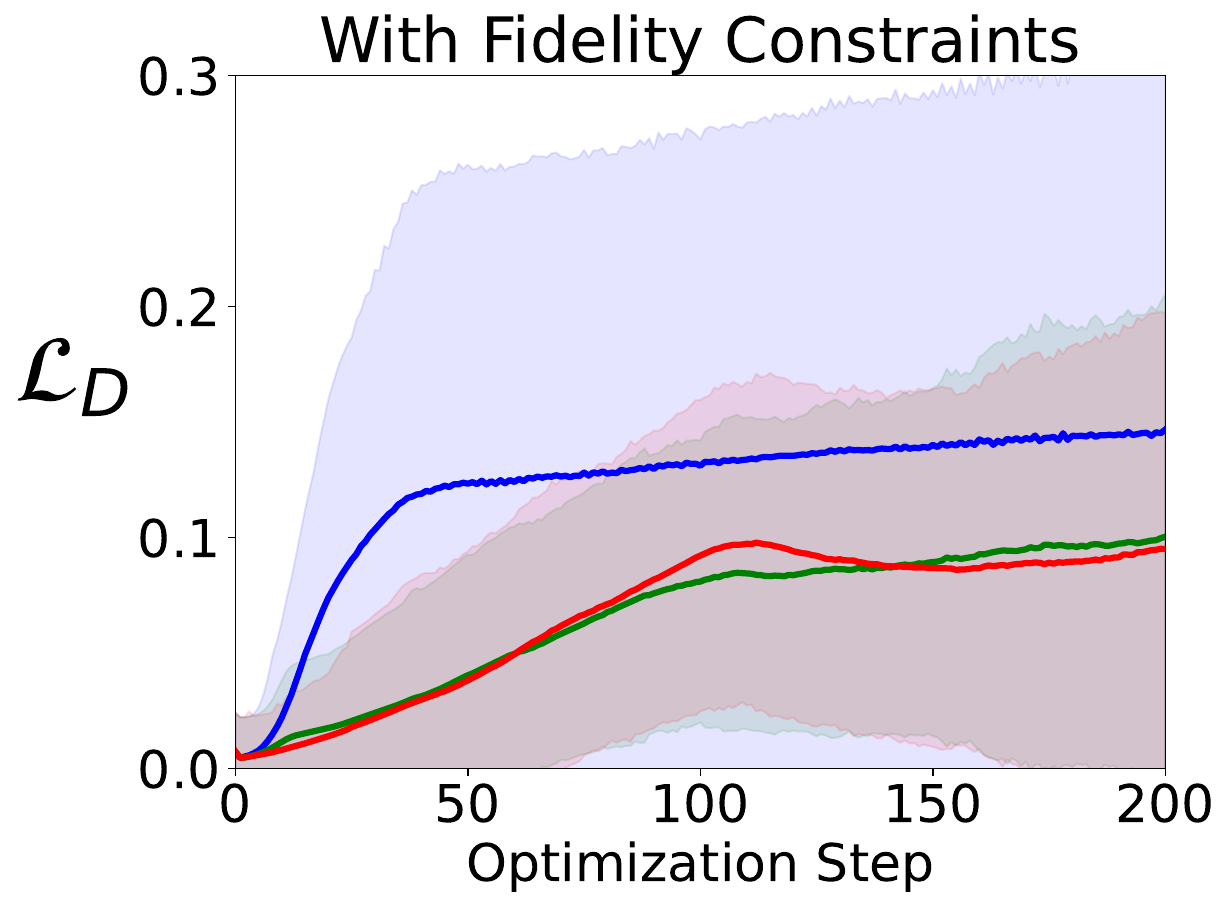}}
  }
  {\caption{Adult Dataset. Distance to closest training sample as a function of gradient steps.}\label{fig:ld}}
\end{figure}

\begin{figure}[t]
\centering
 {%
    \includegraphics[width=0.23\linewidth, height=0.2\linewidth]{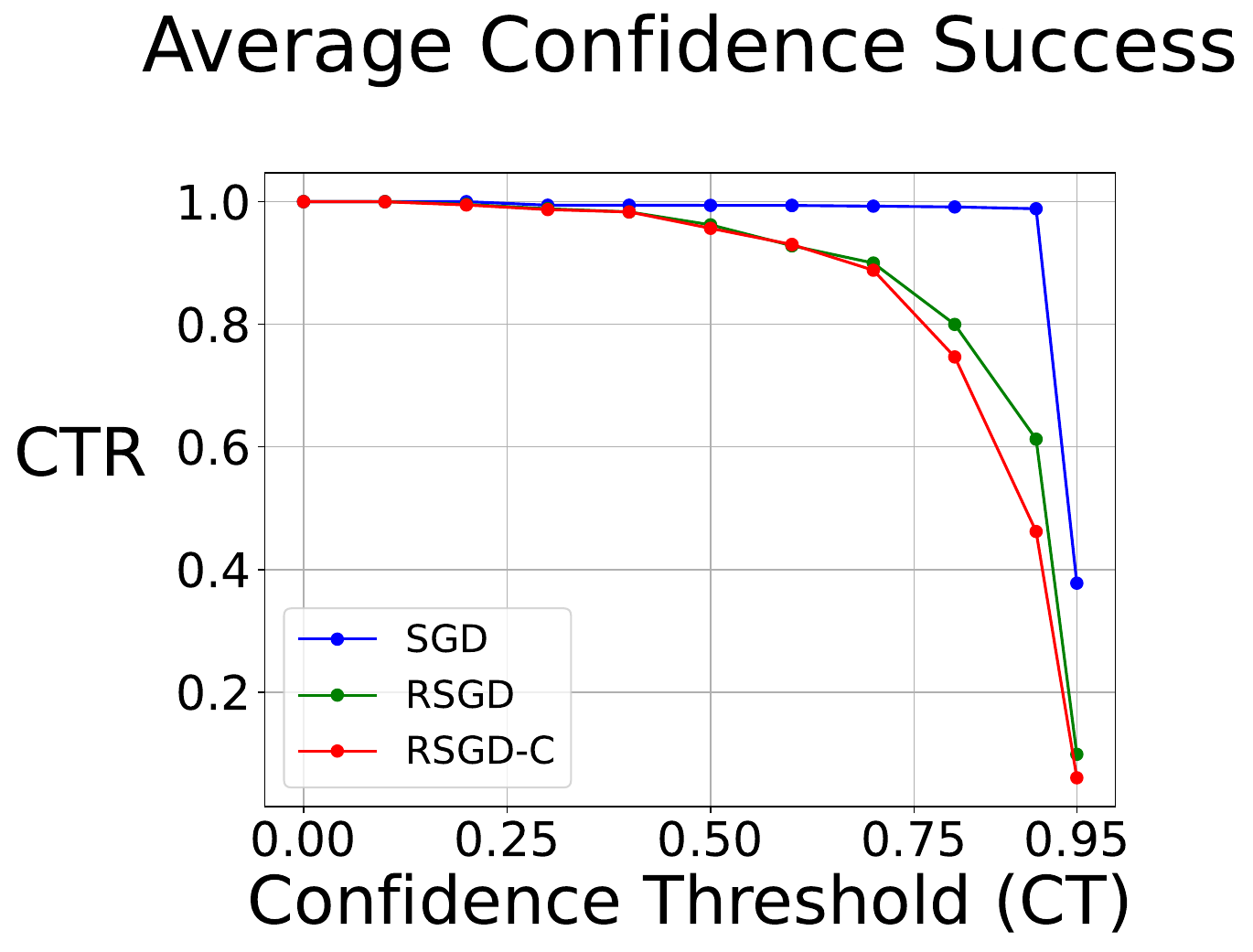}%
    \includegraphics[width=0.23\linewidth, height=0.2\linewidth]{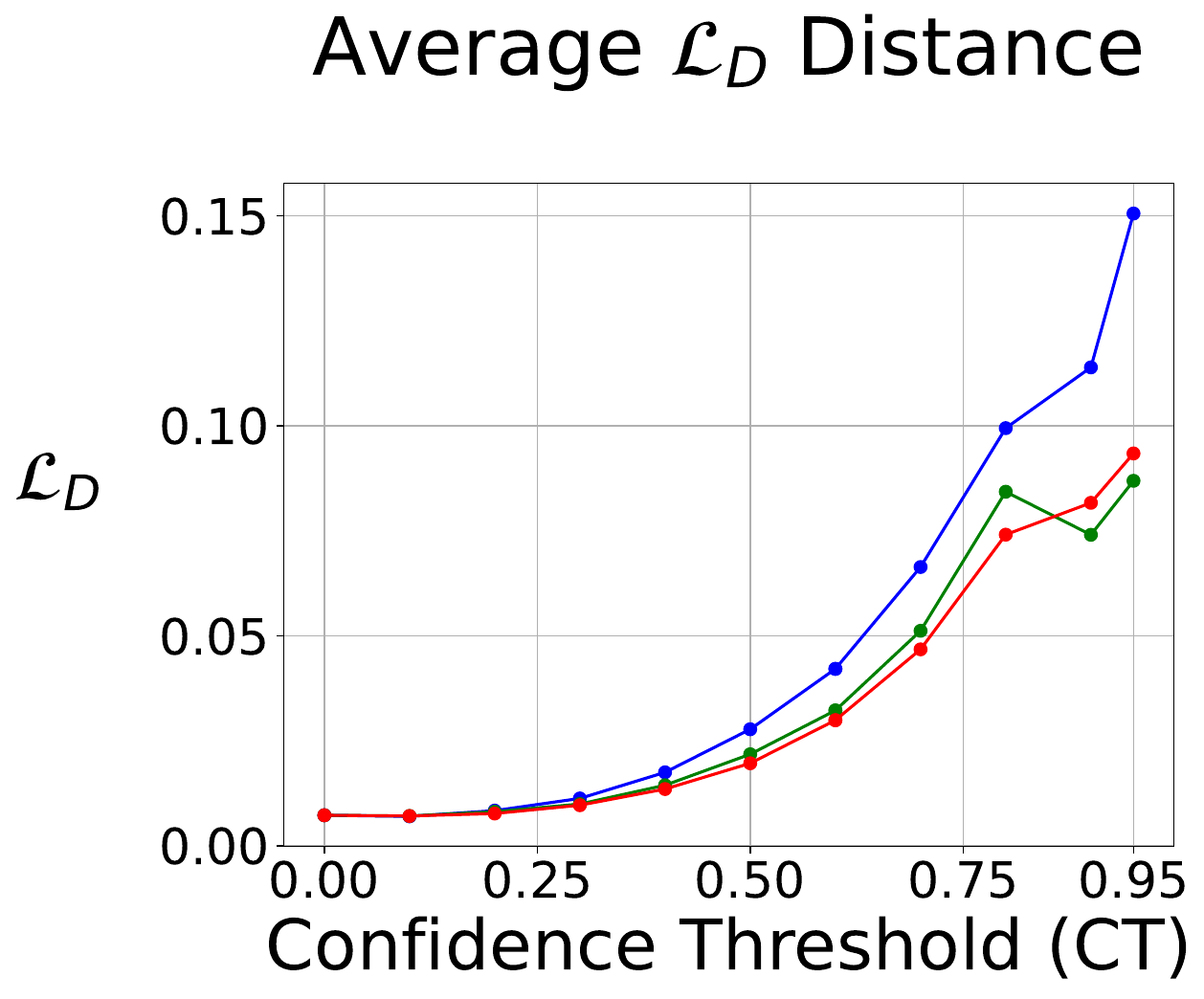}
     \includegraphics[width=0.23\linewidth, height=0.2\linewidth]{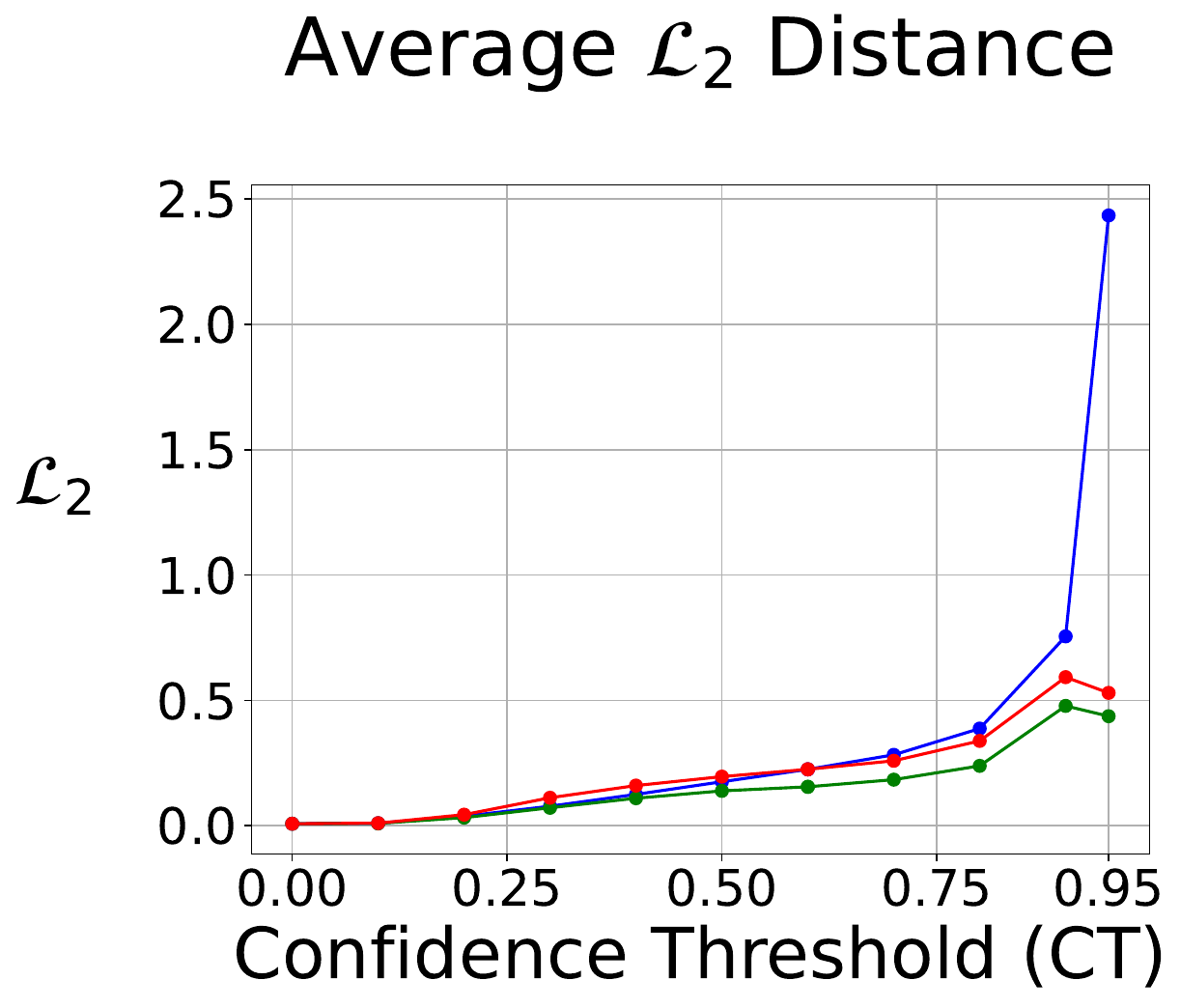}%
    \includegraphics[width=0.23\linewidth, height=0.2\linewidth]{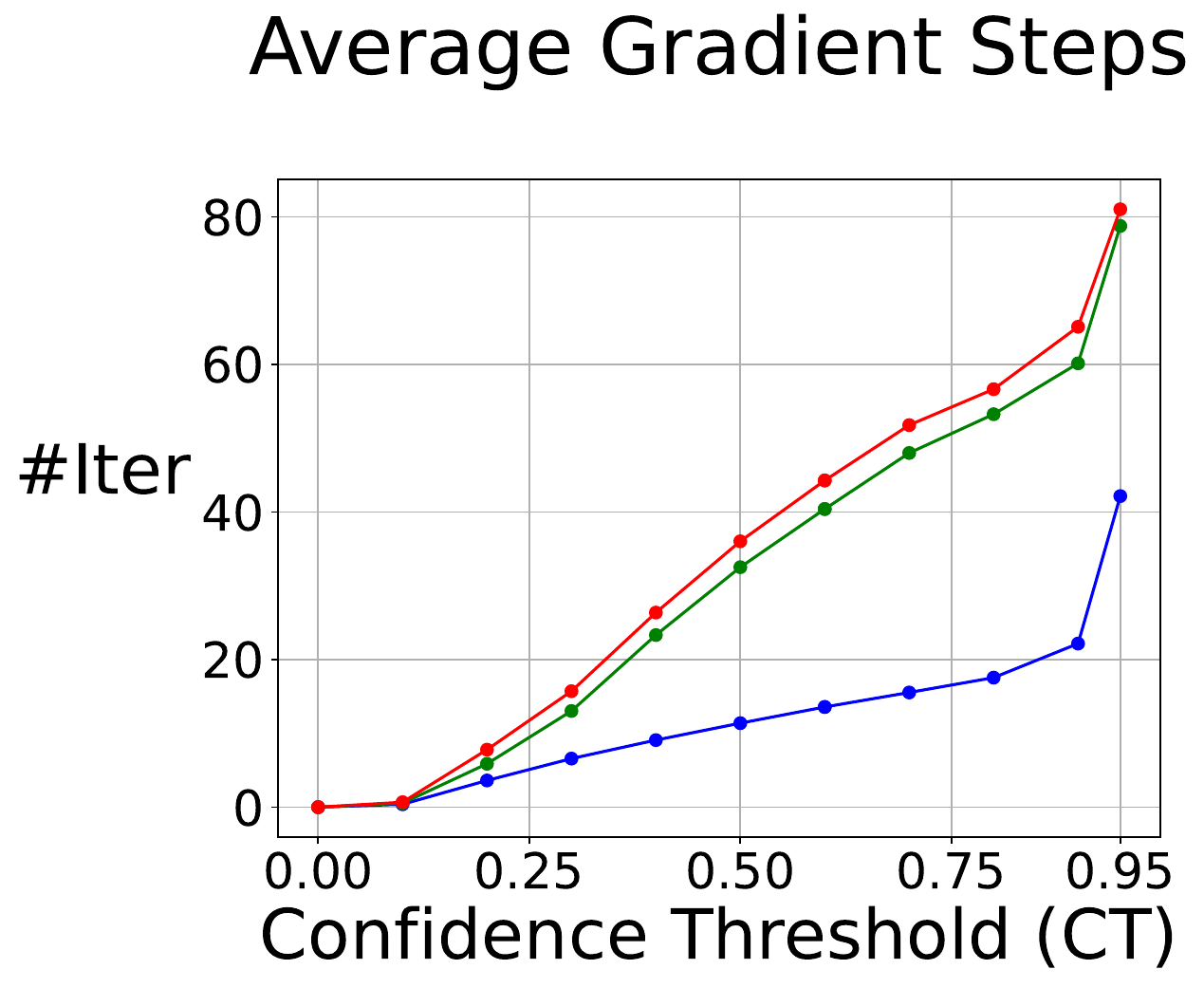}
  }
   {\caption{Adult Dataset. Evaluation of CEs as a function of confidence threshold (CT). Confidence Threshold Ratio (CRT) quantifies 
   confidence success, i.e., the proportion of CEs achieving a specified CT. $\mathcal{L}_D$ measures fidelity to the data manifold (realism), $\mathcal{L}_2$ the fidelity to the initial data point (closeness), and $\#$Iter the number of required gradient steps to achieve a specific confidence level.
   }\label{fig:confidence}}
\end{figure}

\subsubsection{Experimental Setup}
We train a VAE with $d=5$, and a classifier with a representation space of $H=24$ to pull geometry back to the latent space, for both datasets. Implementation details can be found in the Appendix. 
We evaluate different optimizers for generating CEs by traversing the latent space of the VAE, using~\equationref{eq:gaussian_pullback}, where standard SGD uses an Euclidean metric with $\b{M}_\Z^{-1}(\b{z})=\Id$, and setting $\eta = 0.1$ for all methods. To standardize comparison, we fix the number of iterations across methods. Following~\cite{pawelczyk1carla}, we generate CEs for correct negative predictions,  resulting in 7859 samples from 9268 negatives in the Adult test set and 1220 from 1913 in the GMC test set, with the target label set to the positive class. \cite{joshi2019towards} introduce fidelity constraints to keep CEs close to the original input $\b{x}$ during optimization in the VAE’s latent space, using the counterfactual loss: $\ell = \text{BCE}(c(\hat{g}(\b{z}))) + \alpha \| \hat{g}(\b{z}) - \b{x} \|_2$, where BCE is binary cross entropy and $\alpha$ regularization parameter. However, our Riemannian optimizers, namely \RSGD{} and \RSGDCLS{}, naturally respect the data's topology, eliminating the need for such constraints. To showcase this, we evaluate all optimizers both with ($\alpha = 0.1$) and without ($\alpha = 0$) fidelity regularization.

\subsubsection{Results}
\tableref{tab:results-gmc} lists results for Adult dataset, with~\figureref{fig:ld} illustrating local Euclidean distance $\mathcal{L}_D$ across gradient steps. Both \RSGD{} and \RSGDCLS{} maintain counterfactuals close to the data manifold ($\mathcal{L}_D$) and the original input ($\mathcal{L}_0$, $\mathcal{L}_1$, $\mathcal{L}_2$, $\mathcal{L}_{\infty}$) without requiring fidelity constraints during optimization. After 100 iterations, our optimizers generate CEs closer to the original input
compared to \SGD{} with fidelity constraints and 50 steps while maintaining comparable validity and fewer violations of immutable features.  As expected our Riemannian approach requires more gradient steps to achieve high-confidence CEs. To compare optimizers across varying confidence thresholds (CTs), we assess the proportion of CEs successfully generated at each threshold, measured by the Confidence Threshold Ratio (CTR), by selecting the first CE reaching a CT from the counterfactual trajectories. \figureref{fig:confidence} illustrates the average required gradient steps, as well as the fidelity to both the data manifold and the original input, where our approach demonstrates improved fidelity across different confidence levels. 

Similarly, we provide the results for the GMC dataset in~\tableref{tab:results-gmc} and~\figureref{fig:confidence2} in the Appendix. \RSGD{} and \RSGDCLS{} consistently generate high-fidelity CEs but with the cost of less confident CEs compared to \SGD{}. At the same time, \RSGDCLS{} pushes CEs further from an initial factual point to achieve CEs at the same confidence level compared to \RSGD{}, but still it maintains them close to the data manifold, which could potentially be explained by different counterfactual trajectories guided by the geometry of the classifier.

\section{Discussion}

We generate CEs via Riemannian latent space traversal by employing the Riemannian metric induced by the stochastic generator of a deep generative model. Using \RSGD{} we perform CE optimization in the latent space of a VAE, where the metric acts as a preconditioner that keeps the trajectory near the latent data manifold. In addition, we explore a classifier-guided enhanced latent space Riemannian metric, which pulls back the Riemannian metric from the ambient space and captures the geometry of the learned hidden representation manifold induced by the classifier under observation, yielding our \RSGDCLS{} optimizer. 

Different from \SGD{} in the latent space of a Normalizing Flow~\citep{dombrowski2021diffeomorphic}, we ensure that the CE trajectories stay on the data manifold without crossing unpopulated regions in the data (\figureref{fig:model-example,fig:flow}). Our experiments in real-world datasets demonstrate that our approach results in CEs that indeed stay closer to the data manifold (\figureref{fig:ld}) compared to \SGD{}, giving high fidelity with respect to the original factual input and comparable validity (\tableref{tab:results-gmc} and \figureref{fig:confidence}). In general, \RSGD{} and \RSGDCLS{} perform similarly but for the GMC dataset, we observe that trajectories for \RSGDCLS{} are pushed further from the original input compared to \RSGD{} to achieve comparable confidence levels (\figureref{fig:confidence2}). However, as \RSGDCLS{} is guided by the classifier under observation, we hypothesize that these counterfactual trajectories might represent more realistic and actionable paths. This is also illustrated by case 4 in~\figureref{fig:model-example} where \RSGD{} and \RSGDCLS{} generate very different paths, with \RSGD{} running more smoothly through the data.

\paragraph{Societal impact.} Counterfactual explanations are widely used to analyze classifier behavior. A method that generates more realistic counterfactuals can positively impact society by providing debugging tools and promoting fairness. However, like many AI models, this approach could also be misused for malicious purposes.

\paragraph{Limitations and outlook.} We rely on a well-calibrated stochastic generator to approximate the data and provide reliable uncertainty estimates.
Without this, the Riemannian metric in the latent space cannot effectively capture the geometry of the data manifold. While VAEs help induce Riemannian geometry in the latent space, existing methods~\citep{detlefsen:neurips:2019, arvanitidis:iclr:2018} for handling uncertainty are limited to low-dimensional spaces.
For relatively higher-dimensional data, decoder ensembling~\citep{syrota2024decoder} or Laplacian Autoencoders~\citep{miani2022laplacian} may be more suitable. Even though frameworks, e.g.,  Stochman~\citep{software:stochman} support Jacobian computation, the computational complexity increases for complex architectures. Adapting our method to modern models like Diffusion Models~\citep{dhariwal2021diffusion}, where understanding Riemannian latent space geometry~\citep{park2023understanding} or developing Riemannian Diffusion Models~\citep{huang2022riemannian} is an open area of research, remains a topic for future work.

\newpage

\acks{Work on this project was partially funded by the Pioneer Centre for AI (DNRF grant nr P1), the DIREC project EXPLAIN-ME (9142-00001B), and the Novo Nordisk Foundation through the Center for Basic Machine Learning Research in Life Science (NNF20OC0062606). The funding agencies had no influence on the writing of the manuscript nor on the decision to submit it for publication.}

\bibliography{paper}

\newpage
\appendix

\section{Results for GMC dataset}\label{apd:resuls-gmc}

\begin{table}[htbp]
    \centering
    \scriptsize
    
\resizebox{\linewidth}{!}{
\begin{tabular}{c|c|c||ccccc|cc} \toprule
        \#Iter & Constraints  & Optimizer & $\mathcal{L}_{D}  \downarrow$ & $\mathcal{L}_{1} \downarrow$  & $\mathcal{L}_{2} \downarrow$ &  $\mathcal{L}_{\infty}  \downarrow $ & $c(\b{x}_{\CE}) \uparrow$ & FR  $\uparrow$& violation $\downarrow$\\ \midrule
 &  & \SGD & 0.024\tiny{$\pm$0.019} & 1.56\tiny{$\pm$0.45} & 0.59\tiny{$\pm$0.32} & 0.56\tiny{$\pm$0.22} & \textbf{0.90\tiny{$\pm$0.07}} & \textbf{1.00} & 0.97\\
50 & \xmark & \RSGD & 0.019\tiny{$\pm$0.014} & 1.21\tiny{$\pm$0.42} & 0.33\tiny{$\pm$0.23} & 0.39\tiny{$\pm$0.17} & 0.78\tiny{$\pm$0.25} & 0.89 & \textbf{0.96}\\ 
 &  & \RSGDCLS & \textbf{0.015$\pm$0.012} & \textbf{1.11$\pm$0.40} & \textbf{0.29$\pm$0.23} & \textbf{0.37$\pm$0.17} & 0.69$\pm$0.27 & 0.81 & \textbf{0.96}\\ \cmidrule{2-10}

 &  & \SGD & 0.023\tiny{$\pm$0.018} & 1.31\tiny{$\pm$0.44} & 0.40\tiny{$\pm$0.26} & 0.44\tiny{$\pm$0.19} & \textbf{0.86\tiny{$\pm$0.16}} & \textbf{0.97} & 0.97\\
50 & \cmark & \RSGD & 0.019\tiny{$\pm$0.015} & 1.10\tiny{$\pm$0.41} & 0.27\tiny{$\pm$0.20} & 0.35\tiny{$\pm$0.15} & 0.77\tiny{$\pm$0.26} & 0.89 & \textbf{0.96}\\ 
 &  & \RSGDCLS & \textbf{0.015$\pm$0.012} & \textbf{0.83$\pm$0.32} & \textbf{0.16$\pm$0.13} & \textbf{0.26$\pm$0.12} & 0.61$\pm$0.34 & 0.71 & \textbf{0.96}\\ \midrule \midrule

 & & \SGD & 0.026\tiny{$\pm$0.020} & 1.79\tiny{$\pm$0.57} & 0.75\tiny{$\pm$0.42} & 0.62\tiny{$\pm$0.23} & \textbf{0.92\tiny{$\pm$0.05}} & \textbf{1.00} & 0.98\\
100 & \xmark & \RSGD & 0.021\tiny{$\pm$0.014} & 1.52\tiny{$\pm$0.52} & 0.55\tiny{$\pm$0.36} & 0.53\tiny{$\pm$0.23} & 0.82\tiny{$\pm$0.23} & 0.92 & \textbf{0.96}\\ 
 &  & \RSGDCLS & \textbf{0.020$\pm$0.014} & \textbf{1.44$\pm$0.52} & \textbf{0.49$\pm$0.35} & \textbf{0.48$\pm$0.22} & 0.79$\pm$0.23 & 0.90 & \textbf{0.96}\\ \cmidrule{2-10}

 &  & \SGD & 0.025\tiny{$\pm$0.018} & 1.43\tiny{$\pm$0.49} & 0.47\tiny{$\pm$0.30} & 0.48\tiny{$\pm$0.20} & \textbf{0.88\tiny{$\pm$0.15}} & \textbf{0.97} & 0.98\\
100 & \cmark & \RSGD & 0.022\tiny{$\pm$0.015} & 1.30\tiny{$\pm$0.48} & 0.39\tiny{$\pm$0.27} & 0.43\tiny{$\pm$0.19} & 0.81\tiny{$\pm$0.24} & 0.90 & \textbf{0.96}\\ 
 &  & \RSGDCLS & \textbf{0.020$\pm$0.016} & \textbf{1.08$\pm$0.39} & \textbf{0.27$\pm$0.21} & \textbf{0.36$\pm$0.17} & 0.73$\pm$0.29 & 0.83 & \textbf{0.96}\\ \midrule \midrule

 & & \SGD & 0.026\tiny{$\pm$0.020} & 1.83\tiny{$\pm$0.58} & 0.77\tiny{$\pm$0.43} & 0.63\tiny{$\pm$0.23} & \textbf{0.92\tiny{$\pm$0.04}} & \textbf{1.00} & 0.99\\
150 & \xmark & \RSGD & 0.024\tiny{$\pm$0.015} & 1.62\tiny{$\pm$0.54} & 0.62\tiny{$\pm$0.40} & 0.55\tiny{$\pm$0.23} & 0.83\tiny{$\pm$0.21} & 0.93 & \textbf{0.98}\\ 
 &  & \RSGDCLS & \textbf{0.022$\pm$0.014} & \textbf{1.56$\pm$0.55} & \textbf{0.57$\pm$0.39} & \textbf{0.52$\pm$0.23} & 0.80$\pm$0.23 & 0.91 & \textbf{0.98}\\ \cmidrule{2-10}

 &  & \SGD & 0.026\tiny{$\pm$0.019} & 1.44\tiny{$\pm$0.50} & 0.48\tiny{$\pm$0.30} & 0.49\tiny{$\pm$0.20} & \textbf{0.88\tiny{$\pm$0.15}} & \textbf{0.97} & 0.98\\
150 & \cmark & \RSGD & 0.025\tiny{$\pm$0.015} & 1.36\tiny{$\pm$0.49} & 0.43\tiny{$\pm$0.30} & 0.46\tiny{$\pm$0.19} & 0.82\tiny{$\pm$0.23} & 0.91 & \textbf{0.96}\\ 
 & & \RSGDCLS & \textbf{0.021$\pm$0.015} & \textbf{1.18$\pm$0.42} & \textbf{0.33$\pm$0.25} & \textbf{0.39$\pm$0.18} & 0.76$\pm$0.26 & 0.87 & \textbf{0.96}\\ \bottomrule

    \end{tabular}
    }
    \caption{GMC dataset. To standardize comparison, we evaluate optimizers with (\cmark) and without (\xmark) fidelity constraints \emph{during} optimization for different numbers of steps.}
    \label{tab:results-gmc}
\end{table}

\begin{figure}[htbp]
\centering
 
  {%
    \includegraphics[width=0.23\linewidth, height=0.2\linewidth]{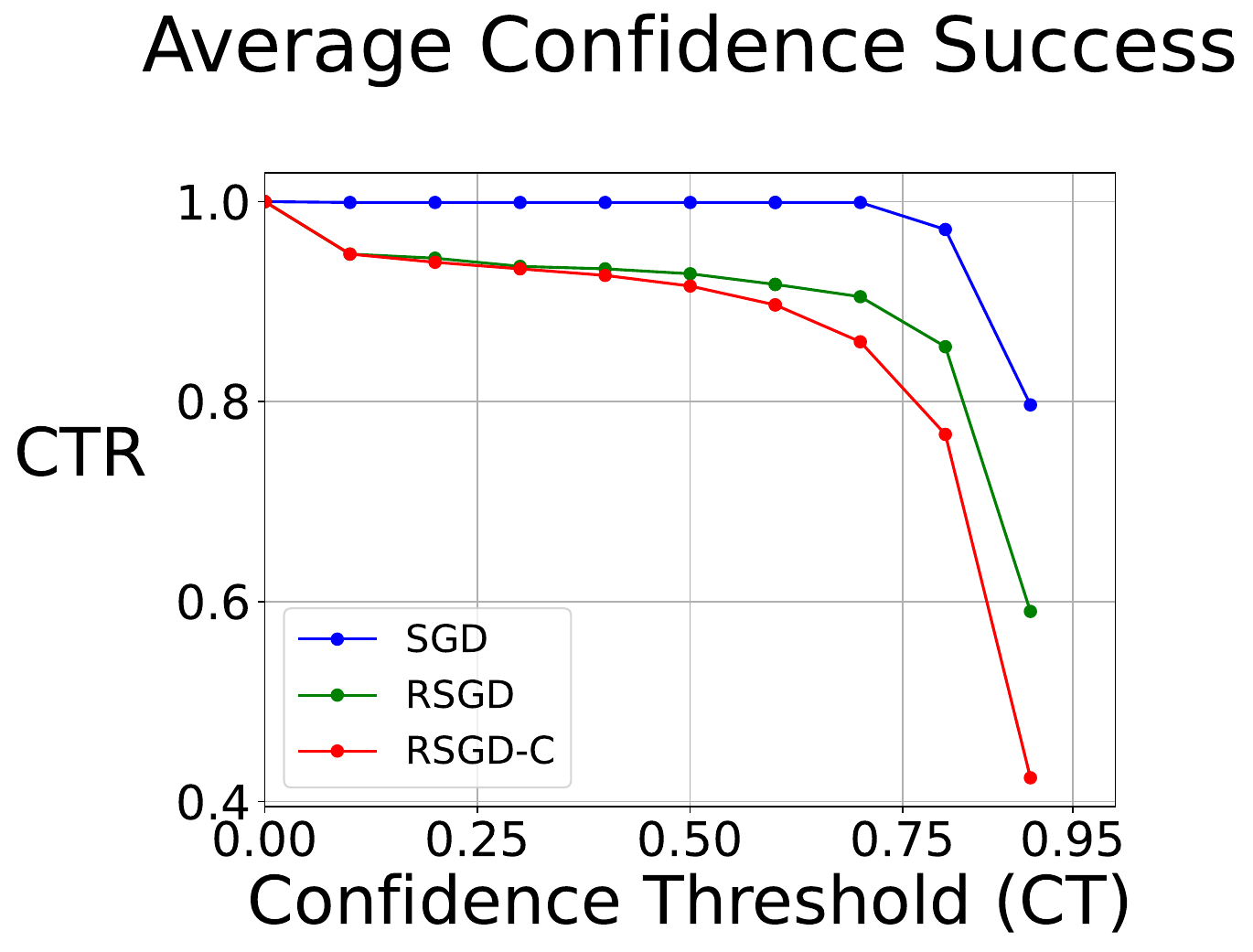}%
    \includegraphics[width=0.23\linewidth, height=0.2\linewidth]{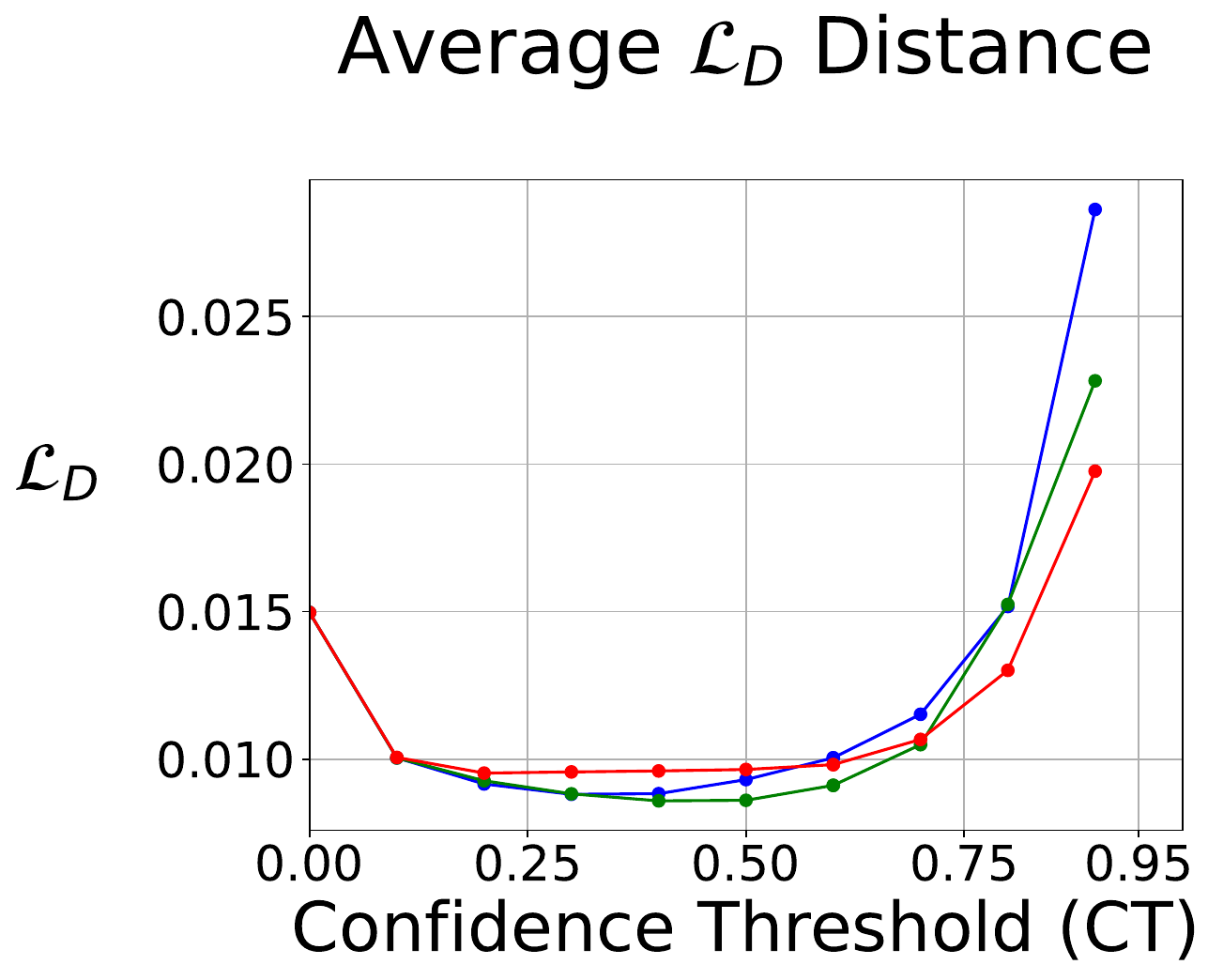}
     \includegraphics[width=0.23\linewidth, height=0.2\linewidth]{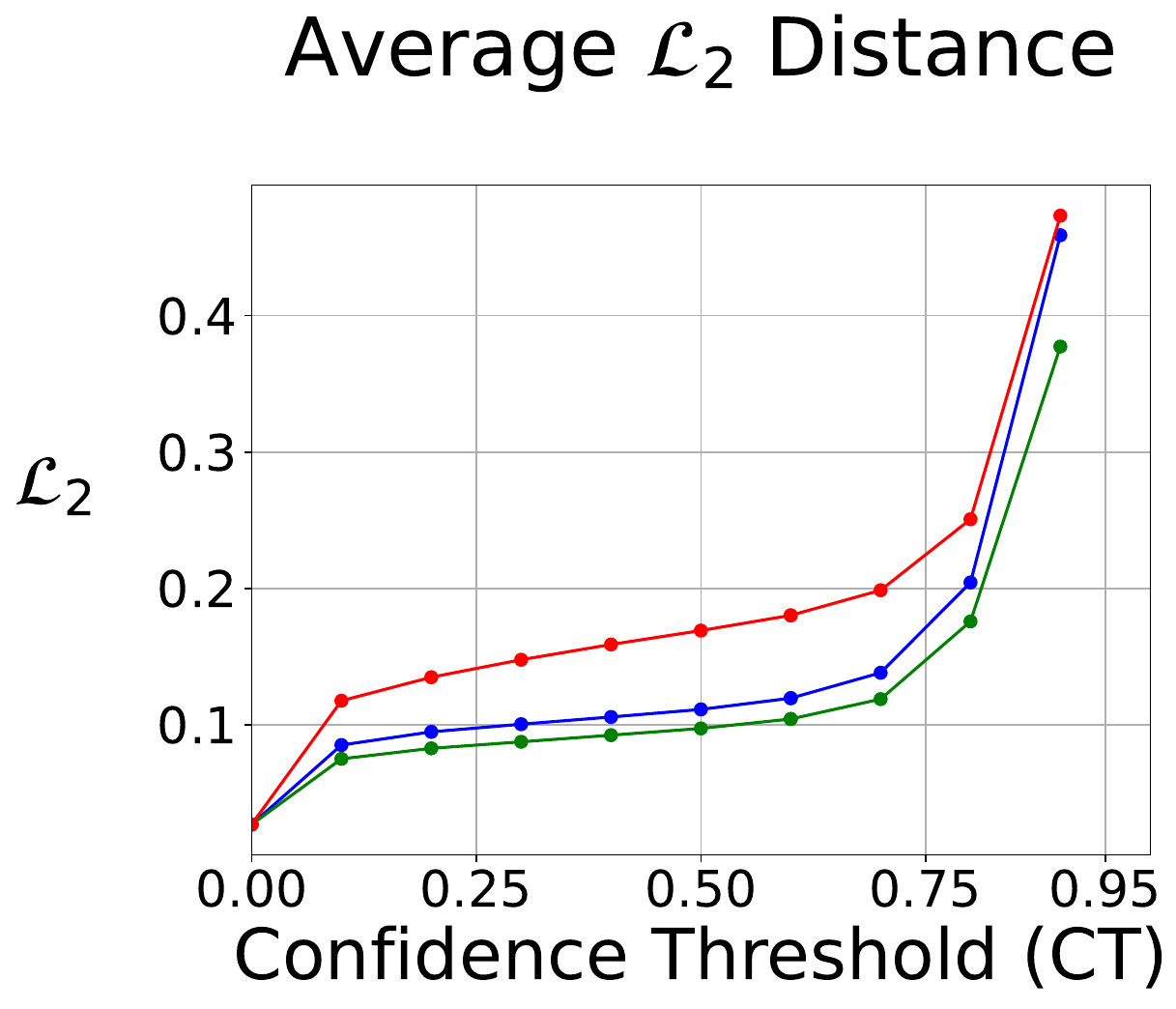}%
    \includegraphics[width=0.23\linewidth, height=0.2\linewidth]{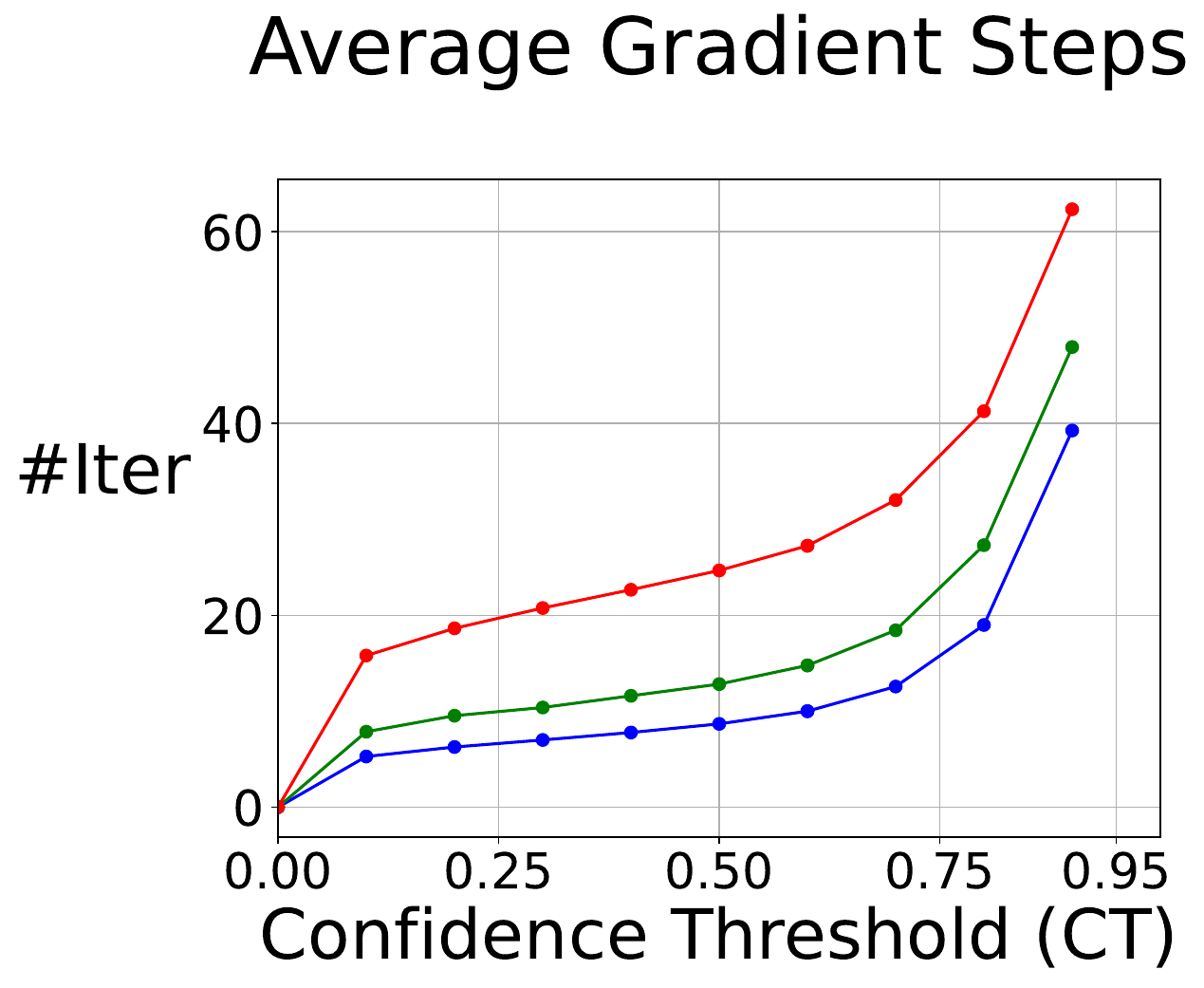}
  }
   {\caption{GMC Dataset. Evaluation of CEs as a function of confidence threshold (CT). Confidence Threshold Ratio (CRT) quantifies 
   confidence success, i.e., the proportion of CEs achieving a specified CT. $\mathcal{L}_D$ measures fidelity to the data manifold (realism), $\mathcal{L}_2$ the fidelity to the initial data point (closeness), and $\#$Iter the number of required gradient steps to achieve a specific confidence level.
   }\label{fig:confidence2}}
\end{figure}

\section{Implementation Details}

\subsection{Datasets}
\textbf{Adult}: The dataset includes $\sim49$K census records with $D=13$ features. The goal is to predict whether an individual's income exceeds \$50K/year ($y=1$, 24\% of samples). Following~\cite{pawelczyk1carla}, we binarize categorical features as aggregated versions of original categories indicating whether the individual works in \textit{private industry}, is \textit{not married}, has an \textit{``other" occupation}, is \textit{not a husband}, identifies as \textit{white} (race), \textit{male} (sex), and is \textit{USA native}. The continuous features are \textit{age, final weight, years of education, weekly work hours,} as well as \textit{capital gain and loss} which we log-transform. The input features sex, race, and \textit{age} are considered immutable. 

\textbf{Give Me Some Credit}: We use the processed data by~\cite{pawelczyk1carla}, which includes $\sim116$K samples after removing missing data with $D=10$ continuous features. Furthermore, we log-transform \textit{depht ratio} input feature. The goal is to predict whether an individual will face financial distress within two years ($y=1$, 93\% of samples). The input feature \textit{age} is considered immutable.

We normalize features in $[0,1]$ and split datasets into 75\%/25\% for train/test.


\subsection{Models}
For both datasets, the binary classifier $c$ under observation is a 4-layer MLP with $2 \cdot H$, $2 \cdot H$, $H$, $H$ neurons including Batch Normalization~\citep{ioffe2015batch} layers, and Tanh activation functions before the last layer, where $H = 24$ is the size of representation space. We trained the classifiers with a batch size of 1024, a learning rate of $10^{-5}$, and applied $\ell_{2}$-regularization of $0.05$ for 20 epochs using the RMSprop optimizer, resulting in $77.5\%$ ($77.6\%$) and $73.6\%$ ($73.8\%$) \emph{balanced} test (train) accuracy for Adult and GMC datasets, respectively. 

Our VAEs employ Gaussian generative models with a 3-layer MLP mirrored encoder-decoder, with $D$, 512, 256 neurons with Batch Normalization layers and Tanh activations functions, a latent space of $d=5$, and Sigmoid activation function in decoder's mean output. Following~\cite{detlefsen:neurips:2019} and~\cite{kalatzis2020variational}, we deterministically warmed up the encoders together with the decoders' mean $\mu_{\theta}$ for 100 epochs, with $\beta$-regularization~\citep{higgins2017beta} of $10^{-4}$. Consecutively, we freeze them while optimizing for 300 epochs decoders' uncertainty $\sigma_{\phi}$, parameterized by Radial Basis Function networks~\citep{que2016back}  with a 0.01 bandwidth and 200 and 350 centers for Adult and GMC, respectively. During training, we used a batch size of 512, and the Adam~\citep{kingma2014adam} optimizer, with a learning rate of $10^{-3}$, expect for GMC's $\sigma_{\phi}$ for which we used $10^{-2}$. During inference and CE optimization, we used Stochman~\citep{software:stochman} to track Jacobians through the VAE's decoder and the classifier.

\end{document}